\documentclass[journal]{IEEEtran}

\usepackage{iftex}
\ifPDFTeX
  \usepackage[utf8]{inputenc}
  \usepackage[T1]{fontenc}
\else
  \usepackage{fontspec}
\fi
\usepackage{amsmath,amssymb}
\usepackage{booktabs}
\usepackage{array}
\usepackage{multirow}
\usepackage{graphicx}
\usepackage{xcolor}
\usepackage{tikz}
\usetikzlibrary{arrows.meta,positioning,calc,shapes.misc,decorations.pathreplacing,backgrounds}
\usepackage[american]{circuitikz}
\usepackage{pgfplots}
\pgfplotsset{compat=1.18}
\usepgfplotslibrary{groupplots,statistics}
\usepackage{pgfplotstable}
\usepackage{siunitx}
\usepackage{url}
\usepackage{xurl}%
\usepackage[hidelinks]{hyperref}
\usepackage{cite}
\usepackage{algorithm}
\usepackage{algpseudocode}
\makeatletter
\newcommand{\AlgNote}[1]{\Statex \hskip\ALG@thistlm\hskip 1.6em{\itshape\color{c2} #1}}
\makeatother
\algrenewcommand\algorithmicforall{\textbf{for each}}

\definecolor{c1}{RGB}{31,73,125}    %
\definecolor{c2}{RGB}{110,110,110}  %
\definecolor{c3}{RGB}{160,70,50}    %
\definecolor{c4}{RGB}{140,170,200}  %
\definecolor{c5}{RGB}{190,190,190}  %
\definecolor{cband}{RGB}{225,232,240}
\pgfplotsset{
  paperaxis/.style={
    width=\columnwidth, height=0.62\columnwidth,
    tick label style={font=\footnotesize}, label style={font=\footnotesize},
    legend style={font=\scriptsize, draw=none, fill=white, fill opacity=0.85, text opacity=1},
    title style={font=\footnotesize},
    axis line style={c2!60}, tick style={c2!60},
    grid=major, grid style={c5!50, line width=0.3pt},
    every axis plot/.append style={line width=0.6pt},
  },
}
\tikzset{
  mark0/.style={mark=*, mark size=1.1pt, c1, fill=c1},
  mark1/.style={mark=square*, mark size=1.1pt, c3, fill=c3},
  mark2/.style={mark=triangle*, mark size=1.4pt, c2, fill=c2},
  mark3/.style={mark=diamond*, mark size=1.4pt, c4, fill=c4},
  mark4/.style={mark=pentagon*, mark size=1.3pt, c5!80!black, fill=c5!80!black},
}
\tikzset{
  lever/.style={-{Stealth[length=4pt]}, c1, line width=0.8pt},
  levereff/.style={-{Stealth[length=4pt]}, c3, line width=0.8pt},
  levernum/.style={circle, draw=c1, fill=white, inner sep=0.8pt, font=\scriptsize, text=c1},
  box/.style={draw=c2, rounded corners=1pt, fill=white, font=\footnotesize, align=center, inner sep=3pt},
  boxacc/.style={box, draw=c1, fill=cband},
  arr/.style={-{Stealth[length=4pt]}, c2, line width=0.6pt},
  arracc/.style={-{Stealth[length=4pt]}, c1, line width=0.8pt},
  lab/.style={font=\scriptsize, text=c2},
  ci/.style={c2, line width=0.6pt},
}

\begin{document}

\title{Local AI Agents in Power Electronics Simulation: Deployment, Measurements and Costs}

\author{Yuzhuo~Li, Pasan~Gunawardena, and~Yunwei~Li
\thanks{Manuscript draft, November 2026.}
\thanks{The authors are with the ELITEGrid Research Laboratory, Department of Electrical and Computer Engineering, University of Alberta, Edmonton, AB, Canada (e-mail: yuzhuo@ualberta.ca; pasan@ualberta.ca; yunwei.li@ualberta.ca).}}

\markboth{Ongoing project, for preprint only}{Local AI Agents in Power Electronics Simulation}

\maketitle

\begin{abstract}
Agentic artificial intelligence (AI) is emerging in power electronics, where it drives the simulation loop that engineers used to close largely by hand. In this loop, a converter design, or a study of the network it connects to, is built, run, measured and revised until it satisfies its specification. Such an agent can be served from a cloud API or from an inference device on the premises. For organizations whose designs, customer data or export-controlled material cannot leave the site, and whose simulators are licensed there, local deployment is the natural choice. In such a deployment, the agent becomes an electrical load on the organization's own equipment, i.e., every run consumes power and energy on a local device, generates tokens and takes time. However, existing work focuses mainly on what the agent can achieve with cloud-hosted models. What a local deployment costs in power, energy, tokens and time on real simulation tasks, and what its choices trade against one another, has received little quantitative investigation under comparable conditions. To address this gap, a systematic evaluation framework is presented in this paper. It applies locally served models to agentic power electronics tasks, from the converter level to the system level, and measures their power, energy, token and time costs. Based on these measurements, the capability boundary of a local agent on such tasks and the cost per solved task are characterized. The results are expected to provide practical insight for the use of agentic AI in power electronics and the power systems it serves.
\end{abstract}


\section{Introduction}
\label{sec:intro}

\IEEEPARstart{G}{enerative} artificial intelligence (AI) has opened new possibilities for industrial application and academic innovation. However, the stochastic nature of its output limits its use in safety-critical applications. To overcome this limitation, agentic AI built on large language models (LLMs) has been developed. In an agent, the model calls tools, reads the returned results and determines the next action, so that computation and verification beyond its own ability are delegated to dedicated software. Moreover, feedback from a simulator in the loop has been shown to improve the task success rate substantially \cite{R13_spiceassistant}. Such agents were first applied to software engineering \cite{R26_swenergy} and have since spread to other engineering domains (Section~\ref{sec:related}). Power electronics, and the power systems it increasingly shapes, are a natural field of application. Although the field is built on physical hardware, simulation models have become accurate and efficient enough to support much of the design, verification and analysis work over the life cycle of a system. In addition, simulation results can be verified numerically.

\begin{figure*}[!t]
\centering
\includegraphics{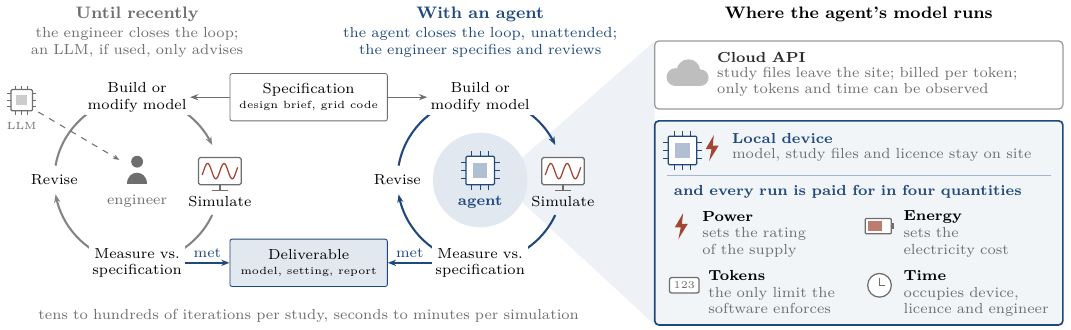}
\caption{The simulation workflow of power electronics and the systems it connects to, and where an agent enters it. A specification is met by building or modifying a simulation model, simulating it, measuring the result against the specification and revising until it is met or shown to be infeasible; the deliverable is a model, a setting or a report. Until recently the engineer closed this loop, with a language model at most advising from outside; an agent closes it itself. The agent's model is served from a cloud API or from a device on the premises, and on a local device every run is paid for in four quantities: power, energy, tokens and time.}
\label{fig:workflow}
\end{figure*}

In power electronics, engineers solve design and study problems by iteration. As shown in Fig.~\ref{fig:workflow}, a converter is modelled, simulated, measured against its specification and revised. A study of the network it connects to runs the same loop over faults, settings and operating points. Until recently, the loop was closed manually, i.e., an LLM used as an advisor remains outside it, and the engineer still runs the simulator, inspects the waveforms and makes the next modification. In contrast, an agent closes the loop itself \cite{R11_pegpt,R12_phia,R14_ietframework,R46_dong_flyback}. Through a tool protocol such as the Model Context Protocol (MCP), it builds or edits the simulation model, or submits simulation requests, reads the measured result and chooses the next action. An agent of this kind needs only a workstation that hosts the simulator and a model to drive it. Therefore, such an agent is within reach of laboratories and of the industrial groups that design drives, converters and protection.

Where the agent runs matters as much as what it can do. It can be served from a cloud API or from an inference device on the premises, and the choice between them is often determined by data-governance requirements before performance is considered. For organizations that handle sensitive designs, customer data or export-controlled material, or whose simulators are restricted to on-site use, local deployment is the natural choice (Section~\ref{sec:related}). Local deployment also makes the agent an electrical load on the organization's own equipment. Every run consumes power and energy on a device the organization owns, generates tokens and occupies a simulator licence for hours. How much a task consumes depends on the choices of the deployment, and neither public leaderboards nor prices give the cost of completing a task. Therefore, four quantities have to be measured, each in its own unit: 1) the power drawn by the inference device, which determines the rating of its power supply; 2) the energy consumed until the task ends, which determines the electricity cost; 3) the number of tokens generated by the model, since the token budget is the only limit enforced by the agent software itself; and 4) the wall-clock time, during which the device, the simulator licence and the engineer awaiting the result are occupied.

The questions that industrial groups ask of such agents are practical, e.g., what the agent can be trusted to do and what it costs to run. The published work reviewed in Section~\ref{sec:related} answers the first of these for cloud-hosted models. However, what a local deployment costs in power, energy, tokens and time on real simulation tasks, and what its choices trade against one another, has received little quantitative investigation under comparable conditions.

To address this gap, a systematic evaluation framework for locally deployed engineering agents is presented in this paper. The main work is summarized as follows: 1) Starting from the structure of the agent loop, expressions for the power, energy, token count and time on an inference device are derived, together with an expression for the expected cost of a task that is run until it succeeds. 2) A unified procedure is built, which tests various agents on the same tasks under identical conditions, checks every result independently of the agent's own report, and records the four quantities for every run. 3) The procedure is applied to tasks that range from the converter level to the system level in simulation environments, and the derived expressions are compared with the measurements.

It should be noted that this study does not attempt an exhaustive evaluation of agents, and its counts are not a ranking of models. Instead, it is an exploration of what a local deployment involves. Its main contribution is not any single figure, but the considerations it identifies and a procedure for evaluating them objectively and fairly. Moreover, the capability of agentic AI is changing quickly. The models measured here will be superseded, and some of the boundaries reported will shift. Therefore, this paper is offered as a first template, i.e., an example that a group considering a local agent can repeat on its own tasks, devices and models.

The rest of this paper is organized as follows. Section~\ref{sec:related} gives the background of local deployment and reviews the related work. Section~\ref{sec:localcloud} analyses the constraints and the mechanism of local deployment, and derives the four quantities on the inference device. Section~\ref{sec:framework} introduces the measurement framework. The measured capability and costs are presented in Section~\ref{sec:cost} and discussed in Section~\ref{sec:discussion}. Finally, Section~\ref{sec:conclusion} concludes this paper.

\section{Background and Related Work}
\label{sec:related}

\begin{table}[!b]
\caption{Prior Work Related to This Paper: What Is Placed in the Loop and What Is Measured}
\label{tab:strands}
\centering\footnotesize
\setlength{\tabcolsep}{3pt}\renewcommand{\arraystretch}{1.08}
\begin{tabular}{@{}>{\raggedright\arraybackslash}p{2.95cm}>{\centering\arraybackslash}p{1.3cm}>{\centering\arraybackslash}p{0.95cm}>{\centering\arraybackslash}p{1.3cm}>{\centering\arraybackslash}p{1.35cm}@{}}
\toprule
Category of work & Simulator in loop & Open weights & Energy measured & Local vs.\ cloud \\
\midrule
Agents for converter design\newline {\scriptsize\cite{R11_pegpt,R12_phia,R13_spiceassistant,R14_ietframework,R45_cui_multiagent,R46_dong_flyback}} & yes & no & no & no \\\addlinespace[2pt]
Agents for other engineering tools and power system solvers\newline {\scriptsize\cite{R6_coscientist,R15_simuagent,R16_feabench,R19_openfoamgpt,R20_poweragentbench_ss,R21_poweragentbench_dyn,R22_psab,R23_gridmind}} & yes & some & no & capability only \\\addlinespace[2pt]
Energy of agents on generic benchmarks\newline {\scriptsize\cite{R24_kaist,R26_swenergy,R31_epg,R40_siraj,R41_yang_armour}} & no & yes & yes & by estimate {\scriptsize\cite{R37_fraunhofer,R51_leitch}} \\\addlinespace[2pt]
Capability indices, open-weight vs.\ closed\newline {\scriptsize\cite{R34_epoch_eci,R35_epoch_lag,R36_epoch_consumer}} & no & yes & no & implied \\
\midrule
This paper & yes & yes & yes, per run & yes, same loop \\
\bottomrule
\end{tabular}
\end{table}

\subsection{Industrial Context and Motivation}
\label{sec:background}

Agentic AI is attracting wide attention and adoption. For example, Gartner \cite{R57_gartner_agentic} expects agentic AI to make at least 15\% of day-to-day work decisions autonomously by 2028, up from none in 2024, and to be built into a third of enterprise software by then. However, Gartner also expects over 40\% of agentic AI projects to be cancelled by 2027 due to escalating cost, unclear value or inadequate risk control.

For an engineering agent, the choice between a cloud API and an inference device on the premises is often determined by data-governance requirements before performance is considered. Organizations that handle sensitive designs, customer data or export-controlled material cannot send their files to a vendor, and an agent reads and writes the engineering files themselves. In line with this, interviews across engineering firms find adoption limited by security and regulatory requirements and legacy toolchains rather than by model capability \cite{R55_edwards_mit}. The simulation software is a second consideration. Simulators are licensed on premises. Some installations can be reached remotely, while others are restricted to on-site use. Where the connection policy of an organization is strict, the model has to be deployed locally. For such organizations, local deployment is the natural choice, and the demand is likely to grow. Two observations support this. In a 2026 survey of 1,000 senior executives, 68\% said that meeting data residency and sovereignty requirements across geographies is challenging \cite{R58_ibm_ibv}. Moreover, the interest in small open-weight models is driven in part by the prospect of running agents on hardware an organization controls \cite{R26_swenergy}.

Local deployment of such an agent is a multidisciplinary undertaking. The power electronics, the data, the algorithms and the simulation tasks are deeply coupled, and the deployment has to bring all of them together. A local run that operates for forty minutes at constant power and fails has consumed energy and time that are not recorded in any bill. How much a task consumes depends on the choices that follow once the agent is local, i.e., which model and quantization \cite{R41_yang_armour,R26_swenergy}, what output-token budget and how much reasoning effort \cite{R24_kaist}, whether measured results are fed back to the agent \cite{R13_spiceassistant}, how many inference nodes, and which tasks to keep local. Each choice is made under two uncertainties, i.e., where the capability of a local agent on the organization's own tasks ends, and what each option costs. For the first, public leaderboards \cite{R34_epoch_eci,R35_epoch_lag} do not say whether a model can edit a Simulink model, read a waveform correctly or keep a protected requirement intact. For the second, neither the price per token nor the price of the hardware gives the cost of completing a task, because a failed run has to be repeated and its cost is added to that of the successful one \cite{R31_epg,R40_siraj}.

The questions that industrial groups ask of such agents are practical: 1) what can the agent be trusted to do, and on what evidence; 2) what does it cost to run; 3) what data leaves the site when a cloud model is used; and 4) which simulation-heavy workflows would benefit.

\subsection{Related Work}

Prior work relevant to these questions falls into four categories. What each category places in the loop and what it measures are summarized in Table~\ref{tab:strands}. The first category builds agents for converter design. Coupling a language model to physics-informed surrogates \cite{R11_pegpt,R11b_pegpt_tie,R12_phia}, to SPICE \cite{R13_spiceassistant,R46_dong_flyback} or to a complete design flow \cite{R14_ietframework,R45_cui_multiagent} improves the task success rate substantially, e.g., from 14\% to 46\% and from 25\% to 85\% on 269 switched-mode supply tasks in one study \cite{R13_spiceassistant}. However, all of these run on closed models, and none measures power or energy. A recent tutorial \cite{R38_li_fundamentals} lists cost, energy, latency and robustness among the considerations that prototype-stage AI for power electronics has under-emphasized. It also observes that practitioners have not yet seen the real boundaries of the methods. Moreover, the one Simulink benchmark that includes open-weight models and power-electronic tasks reports scores for a single task family and does not say where the models ran \cite{R39_simuverity}.

The second category applies agents to other engineering tools: chemistry laboratories \cite{R6_coscientist,R7_chemcrow}, circuit design \cite{R1_analogcoder,R4_davidgoliath}, CAE tools \cite{R15_simuagent,R16_feabench,R18_caeharness,R19_openfoamgpt} and power system solvers \cite{R20_poweragentbench_ss,R21_poweragentbench_dyn,R22_psab,R23_gridmind,R47_etingov_psse}. Where open-weight models are compared with closed ones on the same task, their success rates are usually considerably lower \cite{R1_analogcoder,R19_openfoamgpt,R20_poweragentbench_ss,R21_poweragentbench_dyn}, though not always \cite{R15_simuagent,R48_conductor,R49_embedded}. Moreover, the gap depends on the agent software around the model \cite{R56_openjarvis}. The closest work to the power system set here is PowerAgentBench \cite{R20_poweragentbench_ss,R21_poweragentbench_dyn}. It runs phasor-domain steady-state and dynamic studies in PSS/E, PowerFactory and ANDES against undisclosed evaluators. Besides correctness, it scores whether a submitted result is supported by evidence, whether it is unsafe although declared safe, and whether it respects the stated constraints. However, it reports neither tokens nor energy, and it lists electromagnetic-transient studies as future work. Cost in this work appears as tokens, wall-clock time or API charges at most. These studies also differ in how the model is coupled to the software, and the couplings are compared in Section~\ref{sec:attachment}.

The third category measures the energy of agents on generic benchmarks. It has established the scale of the problem, from a 62 to 137-fold rise in energy per query for the heaviest agent designs \cite{R24_kaist} to nine-fold differences between frameworks \cite{R26_swenergy}. In addition, it has defined energy per successful goal \cite{R31_epg}, derived its dependence on success probability from a retry model \cite{R40_siraj}, and measured energy per correct answer across 86 open-weight checkpoints \cite{R41_yang_armour}. However, none of these places an engineering simulator or an independent physical check in the loop. Moreover, the two local-versus-cloud comparisons on a domain task attach energy by estimate \cite{R37_fraunhofer,R51_leitch}. The fourth category comprises the public indices that place the best open-weight models some months behind the closed frontier on coding, mathematics and knowledge benchmarks \cite{R34_epoch_eci,R35_epoch_lag,R36_epoch_consumer,R44_epoch_cyber}. These indices are the figures on which local deployment is currently argued. However, they are neither task-level boundaries nor costs. To our knowledge, no study measures the power or energy of a language-model agent driving MATLAB/Simulink, PSCAD, SPICE or a finite-element tool.

\section{Local Deployment: Constraints and Mechanism}
\label{sec:localcloud}

Based on the structure of the agent loop and the operating characteristics of the inference device, this section explains how the power, energy, tokens and time of a local deployment arise, and what they require of a device. It uses only the loop, the device and published specifications. The measurements that test it are reported in Section~\ref{sec:cost}, and its application to the model and device of this study in Section~\ref{sec:sizing}.

\subsection{Differences Between Cloud and Local Deployment}
\label{sec:deployments}
\label{sec:hard}

A local run and a cloud run differ by design only in where the model executes (Section~\ref{sec:attachment}), and that location determines what can be measured. On a local node, all four quantities can be measured directly. For a cloud model, the token counts and the wall-clock time are always available, whereas power and energy depend on the access the provider grants. Power and energy can be read when an open-weight model runs on rented hardware that reports its power. For a closed model served through an API, they can only be estimated, with an uncertainty that a local measurement does not carry. A cloud model is priced per token, and Section~\ref{sec:cloud} shows that the price per token is not the cost per solved task.

\begin{figure}[!t]
\centering
\includegraphics{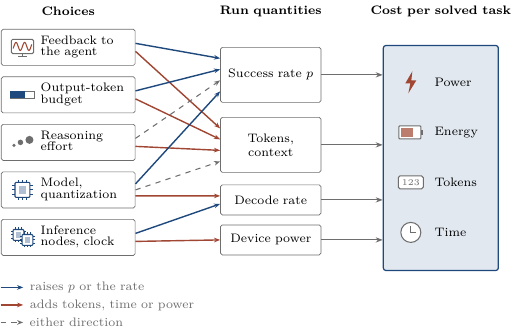}
\caption{The choices of a local deployment, the run quantities each changes, and the costs per solved task they lead to. Most choices change more than one quantity, with opposing effects on the cost, so no choice can be assessed by itself.}
\label{fig:coupling}
\end{figure}

In the cloud, the operator chooses a model and a reasoning tier and pays per token, while capacity, context length and generation rate are handled by the provider. On premises, each of these becomes a choice the operator makes and pays for in time and energy, and four difficulties follow. First, capability: the models a site can serve are smaller than the largest hosted ones. Therefore, on demanding tasks, the success rate $p$ can be low enough that failed runs dominate the cost per solved task (Section~\ref{sec:esuccess}). Second, the device bounds the run. Its memory must hold the model weights and the context together, so a larger model leaves less room for context. Moreover, its memory bandwidth fixes the generation rate, so a long task is a long run at near-constant power. Third, the agent cannot observe its own limits. The output-token budget (cap), chosen by the operator, is enforced by the executor, which stops the run when it is reached. However, this budget is invisible to the model unless the loop reports it (Implication~6 below). Fourth, the choices are coupled. As shown in Fig.~\ref{fig:coupling}, most of them affect more than one quantity with opposing effects on the cost. The coupling that dominates is that of the token budget, which raises $p$ and also the energy of every failed run. The rest of this section derives how each choice acts through the loop, and Section~\ref{sec:cost} measures the couplings one at a time.

\subsection{The Agent Loop}
\label{sec:loop}

\begin{figure*}[!t]
\centering
\includegraphics{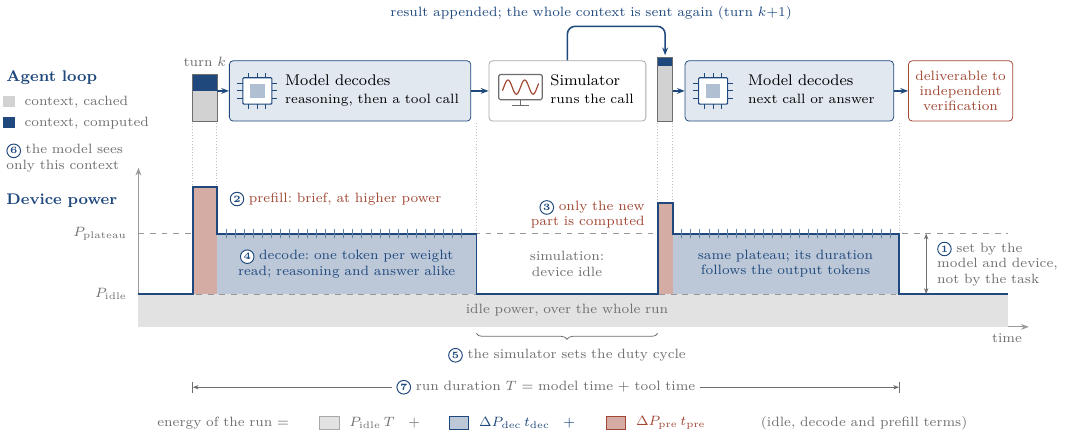}
\caption{The agent loop and the device power it draws, over two turns (sketch, not to scale). Top: on each turn the executor sends the whole context to the model (grey, the prefix already processed and cached; blue, the part computed anew); the model decodes reasoning and a tool call; the simulator runs while the device waits; its result is appended and the whole context is sent again. Bottom: device power on the same time axis, a brief prefill pulse, a decode plateau set by the model and the device rather than the task, and idle power while the simulator runs; the shaded areas are the three terms of \eqref{eq:energy}, written below in the same colours. The model observes only the context it is sent, not the elapsed time, the power or its own token count. Circled numbers mark where each of the seven implications of Section~\ref{sec:mechanism} appears.}
\label{fig:loop}
\label{fig:turnprofile}
\end{figure*}

As shown in Fig.~\ref{fig:loop}, an agent is a language model run in a loop by the executor. The terms used from here on are collected in Table~\ref{tab:terms}. On each turn, the executor assembles a context, consisting of the task text, the descriptions of the tools the model may call and the complete history of the run, and sends it to the model as one request. The model returns output tokens of two kinds, i.e., reasoning, which is intermediate text that the executor does not act on, and a visible message, which is either a structured tool call or a final answer. If it is a tool call, the executor runs the tool, appends the result to the history and sends the whole context again. A run is a sequence of such turns and ends with a final answer, a timeout or a budget stop. The converter's or the network's control still executes inside the simulation at its own time step. The agent acts between simulations, on a timescale of minutes, as a supervisory outer loop around the experiment. Three properties of this loop determine its cost. First, the context grows monotonically. Every turn re-sends everything before it. Second, the model is the only decision-making component. The number of turns in a run, whether two or two hundred, is determined by the model. Third, the model observes only its context. The elapsed time, the power and its own token count are unknown to it unless the loop writes them into the context. These properties hold identically for a local and a cloud model. Three features distinguish an engineering agent from the more widely studied software agent: 1) tool calls are simulations and are slow, so the simulator imposes a duty cycle on the device (Implication~5 below); 2) the deliverable is a model file or a setting, which must be simulated rather than read to be verified; and 3) the task has a numerical ground truth, so the result can be checked at operating points the agent did not see.

\begin{table}[!t]
\caption{Terms Used in This Paper}
\label{tab:terms}
\centering\footnotesize
\begin{tabular}{@{}p{1.9cm}p{6.1cm}@{}}\toprule
Term & Meaning here \\ \midrule
Executor & The agent software that runs the loop: assembles the context, calls the model, runs the tools and enforces the token budget. \\
Context & Everything sent to the model on one request: the task text, the tool descriptions and the full history of the run. \\
Prefill, decode & The two phases of a request on the inference device: reading the context (short, high power), then generating output one token at a time (long, near-constant power). \\
Output tokens & What the model generates, reasoning and visible message together; the token budget counts these. \\
Reasoning tokens & Output the model produces for itself before its visible message; decoded, and metered, like any other output. \\
Test condition & One task or task family, one model and one configuration, with its repeated runs. \\
Invalid run & A run stopped by an infrastructure or author-side fault; re-run, not counted. \\
$p$ & Success rate: passes over valid runs of a test condition. \\ \bottomrule
\end{tabular}
\end{table}

\subsection{The Four Quantities on the Inference Device}
\label{sec:mechanism}

A request has two phases on the inference device, and electrically they are two operating states of the same load. In the prefill phase, the device processes the input context. All input tokens are handled together, so the computation is dense, fully loads the arithmetic units and is short: a brief pulse at the peak power of the device. In the decode phase, the device produces output tokens one at a time, and for each token it reads the model's active weights once from memory: a long interval at a lower, nearly constant power. A single agent decodes at batch size one, and memory bandwidth sets an upper limit on this phase, i.e., the token rate is at most the bandwidth divided by the bytes of weight read per token. A mixture-of-experts model reads only its active experts. Therefore, a model with a large total parameter count but a small active count can decode at a practical rate with moderate memory bandwidth. Seven implications follow for the four quantities, and Section~\ref{sec:cost} tests each of them. Fig.~\ref{fig:turnprofile} sketches the power the device draws over two turns of the loop, from which the implications can be read.
\label{sec:phases}

\textit{Implication 1.} While decoding, the device does the same work for every token, reading the same quantity of weights and performing the same arithmetic, regardless of the token content. Decode power is therefore a plateau, a constant power level that does not depend on the task, the token budget or the phase of the run; it can differ between models, and it varies with the serving settings \cite{R72_powertrace}. The same bound has been observed on data-centre GPUs \cite{R42_ma_capping}.

\textit{Implication 2.} Prefill draws more power than decode but occupies a small share of the run, because the context of turn $k$ is a prefix of the context of turn $k+1$ and the inference server keeps the computed state of that prefix.

\textit{Implication 3.} For the same reason, the nominal count of input tokens overstates the computation by a large factor, and an energy estimate based on nominal token counts is too high by the same factor.

\textit{Implication 4.} Reasoning tokens, produced when reasoning is enabled, before the visible message, are decoded exactly like other output and counted against the same budget. With reasoning enabled they make up a large share of the output tokens, and therefore of decode time and energy. Disabling reasoning does not change the plateau. Instead, it changes the number of turns and, through the growth of the context, the prefill share and the total token count.
\label{sec:reasoning}

\textit{Implication 5.} When the model issues a tool call, the inference device waits. In an engineering agent, the tool is a simulation, and the wait is seconds to minutes at the idle power. The mean power over a run therefore lies below the plateau by the share of time spent waiting for tools, and the energy of a run decomposes as
\label{sec:tooltime}
\begin{equation}
E_\mathrm{run} = P_\mathrm{idle}\,T + \Delta P_\mathrm{dec}\,t_\mathrm{dec} + \Delta P_\mathrm{pre}\,t_\mathrm{pre},
\label{eq:energy}
\end{equation}
where $T$ is the run duration, $t_\mathrm{dec}$ and $t_\mathrm{pre}$ the times in each phase and $\Delta P$ the power above idle in that phase. Prefill is short, and decode time is the output token count divided by a nearly constant rate. Therefore, the energy of a run is set by its duration first and by its output token count second.

\textit{Implication 6.} The executor enforces the output-token budget, the only resource limit counted in the model's own units, and by the third property of the loop the model does not see the count unless the loop writes it into the context. Without a running indication of the tokens consumed, a run can therefore be stopped at the cap when it is close to finishing. Writing the consumption into the tool results can be expected to change how runs end before it changes whether they succeed.
\label{sec:budget}

\textit{Implication 7.} The wall-clock time of a run is the model time plus the tool time, $T \approx N_\mathrm{out}/r + t_\mathrm{pre} + t_\mathrm{tool}$, where $N_\mathrm{out}$ is the output token count and $r$ the decode rate that the device and the active weights fix (Implication~1). In a local deployment, a lower generation rate increases the time per output token, and lower capability increases the number of output tokens per run. The simulator adds $t_\mathrm{tool}$ whichever model runs. Time is also the one cost that can be measured in both a local and a cloud deployment, and the one on which they can be compared directly.
\label{sec:time}

\subsection{Expected Cost per Solved Task}
\label{sec:esuccess}

Let $p$ be the probability that a run passes, $E_\mathrm{pass}$ the mean energy of a passing run and $E_\mathrm{fail}$ the mean energy of a failing run. Over repeated attempts, the expected energy spent per solved task is
\begin{equation}
E_\mathrm{success} = \frac{p\,E_\mathrm{pass} + (1-p)\,E_\mathrm{fail}}{p} = E_\mathrm{pass} + \frac{1-p}{p}\,E_\mathrm{fail}.
\label{eq:esuccess}
\end{equation}
With a token budget in place, most failures are budget stops, so $E_\mathrm{fail}$ is close to the energy of a full-budget run and is not smaller than $E_\mathrm{pass}$. When $p$ is small, the second term dominates and $E_\mathrm{success}$ scales as $E_\mathrm{fail}/p$. A device that decodes twice as fast halves both energies, while a change that doubles $p$ halves the second term. At $p = 0.5$, the two effects are comparable. At $p = 0.05$, the success rate dominates by an order of magnitude. In this sense, the cost per solved task is determined less by the cost of a single run than by the success rate. Taking the energy of a passing run as the base, $E_\mathrm{pass} = 1$\,pu, and writing $\varepsilon = E_\mathrm{fail}/E_\mathrm{pass}$, \eqref{eq:esuccess} becomes $E_\mathrm{success} = 1 + \varepsilon(1-p)/p$\,pu. The multiplier depends only on the success rate and on how much a failure costs relative to a success, and with $\varepsilon \ge 1$ it is at least $1/p$. The same identity holds for the wall-clock time and the output tokens of a run, with $T$ or $N_\mathrm{out}$ in place of $E$. Therefore, a solved task has an expected energy, time and token cost, each given by the same expression in $p$. Power is the exception. It is a rate, whose scale is the decode plateau of Implication~1, and it enters the cost of a solved task only through the energy. The expression assumes that a failed run is recognized as one, by an independent check of the kind described in Section~\ref{sec:framework}. An incorrect result that is accepted is not covered by it.

\subsection{Device Sizing}
\label{sec:deploy}

The device of this study was chosen as the minimum on which the study could be carried out, not as a deployment. An organization that deploys a local agent chooses the device instead. Three requirements of the task set the choice: 1) the memory that must hold the model and its context; 2) the generation rate that the required completion time demands; and 3) the power and energy the device then draws.

\emph{Capacity.} The memory of the device has to hold three things at once: 1) the model itself, i.e., its parameters (the weights) stored at the chosen numerical precision; 2) a buffer that the model maintains for every token of the context, the key--value cache, which grows in proportion to the history of the run; and 3) the serving software. For a model of $N_\mathrm{tot}$ parameters stored at $b_w$ bytes each and a context of $L$ tokens that occupies $k$ bytes per token,
\begin{equation}
M \;\ge\; N_\mathrm{tot}\,b_w \;+\; k\,L \;+\; M_0 .
\label{eq:capacity}
\end{equation}
The first term is fixed once the model and its precision are chosen, while the second grows as the run proceeds, up to the context window. Capacity is a hard limit of the device, i.e., a model that does not fit cannot be served, and a context that exceeds the memory remaining after the weights are loaded cannot be processed. Not every parameter need be stored at the same precision, and some models carry large look-up tables that are stored but consulted only occasionally. Each part enters the first term at its own size. For a large model, the first term alone approaches the memory of the device. In this case, the choice of model fixes the class of device before any task is run, and the length of history the agent can carry is limited by what remains.

\emph{Decode rate.} A single agent decodes at batch size one, and each output token reads the active weights once (Section~\ref{sec:mechanism} above). With memory bandwidth $B$, $N_\mathrm{act}$ active parameters and an attained fraction $\eta$ of the bandwidth,
\begin{equation}
r_\mathrm{dec} \;\le\; \eta\,\frac{B}{N_\mathrm{act}\,b_w}.
\label{eq:rdec}
\end{equation}
The bound is an upper limit. The attained fraction $\eta$ depends on the model, its routing and the serving engine, and has to be measured for the combination actually deployed. The values attained in this study are given in Section~\ref{sec:sizing}.

\emph{Required rate.} On the local device, the model time dominates the run (Section~\ref{sec:timeres}). Therefore, a task whose solution needs $N_\mathrm{out}$ output tokens and must be returned within a required completion time $T$ needs approximately
\begin{equation}
r_\mathrm{req} \;\approx\; \frac{N_\mathrm{out}}{\beta\,T},
\label{eq:rreq}
\end{equation}
where $\beta$ is the share of the window in which the device is busy rather than waiting for a tool. Equation~\eqref{eq:esuccess} enters on top of this. The completion time that matters is that of a solved task, so the required rate scales with $1/p$ as well.

\emph{Prefill and the context that is re-sent.} An agent re-sends its whole context on every turn, and almost all of it is a prefix that the server has already processed (Implications~2 and~3). A serving engine without prefix caching would recompute the full context on every turn, which at the cache hit rates of an agent run multiplies the prefill work many times over. Therefore, prefix caching is a requirement of an agent deployment rather than an optimization of it. For the same reason, a device's prefill rate at a short prompt says little about an agent run whose context grows to a large part of the context window.

\emph{Operating cost.} Once the device is chosen, the operating cost per solved task follows from \eqref{eq:esuccess} with the device's own constants. The energy per output token during decode is the plateau over the rate, $P_\mathrm{dec}/r_\mathrm{dec}$. A faster device lowers it only if its plateau rises more slowly than its rate, which has to be measured on the device and cannot be taken from its rating. The time per solved task is $N_\mathrm{out}/(\beta\,r_\mathrm{dec}\,p)$ plus the tool time, during which the device, the simulator licence and the engineer awaiting the result are occupied. A device that stays loaded between tasks consumes its idle power in every hour it is not decoding. Purchase prices are not quoted. They change faster than the measurements in this paper, and the quantities that set them, i.e., the capacity of \eqref{eq:capacity} and the bandwidth of \eqref{eq:rdec}, are the ones given here. The resulting sizing procedure is as follows: 1) measure $p$ for the model, the precision and the engine to be deployed on a handful of the organization's own tasks; 2) take the capacity from \eqref{eq:capacity}; 3) take the required rate from the required completion time and \eqref{eq:rreq}; 4) choose the smallest device class of Table~\ref{tab:devices} (Section~\ref{sec:deployclasses}) that meets both; and 5) meter its plateau and idle power on the first runs, because the costs per solved task on any device other than the one measured here remain estimates until then.

\subsection{Device Classes and Memory Tiers}
\label{sec:deployclasses}
\label{sec:deploytiers}

Table~\ref{tab:devices} places the classes of device that can serve a model locally against the two quantities that \eqref{eq:capacity} and \eqref{eq:rdec} ask for, i.e., memory and memory bandwidth, together with the rated power of each. Every entry is a vendor specification, and the devices are examples of their class. The table is not exhaustive and implies no recommendation or preference. Two properties separate the classes. Unified-memory devices and multi-channel server processors offer 128\,GB or more at bandwidths from about 250\,GB/s to 1.2\,TB/s, so they can hold a large mixture-of-experts model whole. On the 128\,GB devices, such a model leaves little margin, and the context competes for what is left. Discrete accelerators offer more bandwidth and less memory per device. The consumer and workstation graphics cards of the table reach about 1.8\,TB/s with 12 to 96\,GB. Therefore, a model of that size either spans several cards or is split between the card and host memory, as described next. Accelerators built for data centres carry more memory at higher bandwidth and a correspondingly higher rated power, usually as rack-scale systems. They are not listed, and the same two relations apply to them. A desktop processor has the capacity at a bandwidth several times lower than any of these. Rated power spans more than an order of magnitude across the classes, and it is the rating, not the decode plateau, that a supply has to be sized for (Section~\ref{sec:power}). Section~\ref{sec:sizing} applies the table to the model of this study.

\begin{table}[!t]
\caption{Device Classes for Local Deployment: Memory, Memory Bandwidth $B$ and Rated Power, from Vendor Specifications. Rated Power Is the Chip TDP for Processors, the Board Power for Graphics Cards and the System Maximum for the Mac Studio}
\label{tab:devices}
\centering
\footnotesize
\setlength{\tabcolsep}{3.5pt}
\begin{tabular}{@{}lrrr@{}}
\toprule
Device class (example) & \begin{tabular}[b]{@{}r@{}}Memory\\(GB)\end{tabular} & \begin{tabular}[b]{@{}r@{}}$B$\\(GB/s)\end{tabular} & \begin{tabular}[b]{@{}r@{}}Rated\\power (W)\end{tabular} \\
\midrule
Desktop CPU, Ryzen 9000 \cite{R68_amd_ryzen9000} & 128$^{a}$ & 90 & 65--170 \\
Unified, Ryzen AI Max+ 395 \cite{R65_amd_ryzenaimax} & 128 & 256 & 45--120 \\
Unified, GB10 \cite{R62_nvidia_dgxspark} & 128 & 273 & 140 \\
Server CPU, EPYC 9005 \cite{R66_amd_epyc} & $\le$6{,}000 & 614 & 450--500 \\
Unified, Apple M5 Ultra \cite{R67_apple_macstudio} & $\le$512 & 1{,}200 & 480 \\
Graphics card, RTX 5070 \cite{R63_nvidia_rtx50} & 12 & 672 & 250 \\
Graphics card, RTX 5090 \cite{R63_nvidia_rtx50} & 32 & 1{,}792 & 575 \\
Workstation, RTX PRO 6000 \cite{R64_nvidia_rtxpro6000} & 96 & 1{,}792 & 600 \\
\bottomrule
\end{tabular}
\par\smallskip\raggedright\scriptsize The devices are examples of each class; the list is not exhaustive and implies no recommendation or preference. $^{a}$Two-channel DDR5-5600 with two DIMMs; four DIMMs (up to 256\,GB) are rated at DDR5-3600, 58\,GB/s.
\end{table}

A mixture-of-experts model does not read all of its weights on every token, and the open-source engines released in 2026 use this to serve models larger than the fastest memory of the device. The weights are placed across tiers: 1) the attention layers, the shared expert and the most frequently routed experts in the graphics card's memory; 2) the remaining experts in host memory; and 3) parameters that are only looked up, such as a large embedding table, in host memory or on solid-state storage. The time per output token is then the sum over tiers of the bytes read from each tier over its bandwidth,
\begin{equation}
\frac{1}{r_\mathrm{dec}} \;\approx\; \sum_i \frac{b_i}{\eta_i B_i},
\label{eq:tiers}
\end{equation}
so the slowest tier that is read on most tokens sets the rate, and keeping the frequently routed experts in fast memory removes bytes from the slow terms. The llama.cpp engine keeps the expert weights of selected layers in host memory and caches frequently routed experts on the card \cite{R60_llamacpp}. KTransformers computes the host-resident experts on the processor with kernels written for its matrix units \cite{R61_ktransformers}. Strata, released in September 2026, serves a large mixture-of-experts model on a single consumer card with the remaining weights in host memory, and reads from solid-state storage the weights that fit in neither \cite{R59_strata}.

These engines lower the capacity a model requires, and with it the class of device. However, they do not by themselves lower the cost per solved task. The rates they report are commonly measured on short answers at short contexts, often at a lower precision and with speculative decoding. Each of these conditions changes a factor of Section~\ref{sec:factors}, and the factor $(1-p)/p$ of \eqref{eq:esuccess} multiplies every per-run saving. Serving several streams at once shares each read of the weights among them and raises the aggregate rate when the contexts are short. However, at the context lengths of an agent run, the key--value cache is read on every token as well (Section~\ref{sec:sizing}).

\begin{figure*}[!t]
\centering
\includegraphics{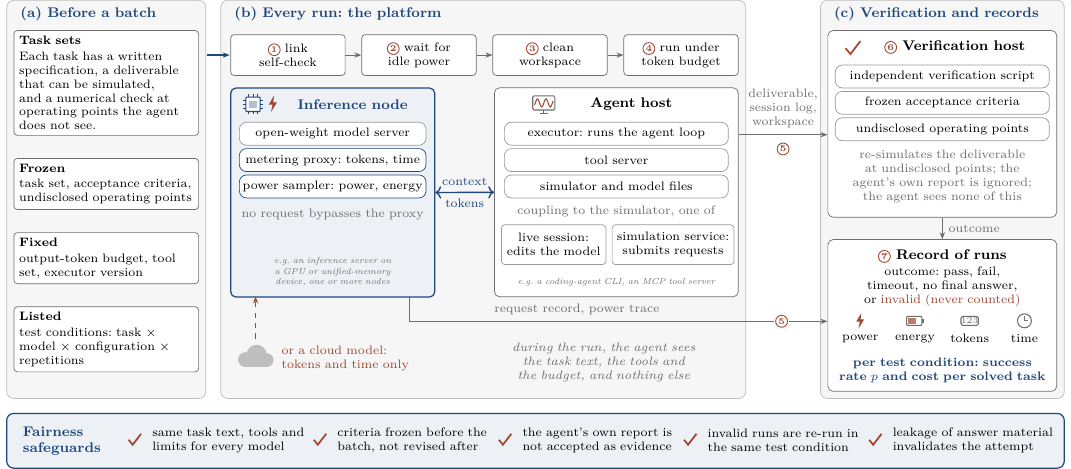}
\caption{The measurement framework. (a) Before a batch, the task sets, the acceptance criteria and the undisclosed operating points are frozen, the output-token budget, the tool set and the executor version are fixed, and the test conditions are listed. (b) Every run takes place on a platform with three roles; the agent host and the inference node are shown with what each provides, and examples of what can fill each role are in italics (Section~\ref{sec:cost} gives this study's). The metering proxy and the power sampler exist only on a local node; a cloud model replaces the node behind the same executor, tools and verification script and reports tokens and time only. (c) The verification host checks the deliverable independently of the agent's own report, and the outcome is recorded with the four quantities. The circled numbers are the steps of the Algorithm, and the safeguards at the bottom apply to every model under test.}
\label{fig:platform}
\end{figure*}

\section{Measurement Framework}
\label{sec:framework}

The procedure has two aims: 1) to determine the capability boundary of an agent on a given set of tasks, objectively and in a way that is fair across models, configurations and deployments; and 2) to record the four quantities for every run from independent sources. Its principles do not depend on a particular simulator, inference device or model, and the same procedure was applied unchanged to both simulator couplings described below. Fig.~\ref{fig:platform} gives an overview of the framework, from what is fixed before a batch to what is recorded for every run.

\subsection{Platform and Coupling to the Simulator}
\label{sec:attachment}

As shown in Fig.~\ref{fig:platform}(b) and (c), the platform has three roles, and the procedure specifies for each only the functions it must provide. Therefore, any executor, simulator, inference device or model that provides them can be substituted. The agent host runs the executor, the tool server and the engineering simulator, and it holds the model files and executes the simulations. The executor is the agent software that runs the loop, and its version is pinned for a batch, because a change of executor is a change of configuration (Section~\ref{sec:factors}). The inference node serves the open-weight model and carries a metering proxy and a power sampler. Every request passes through the proxy, which records its time and token counts, and the model server accepts connections only through the proxy, so that no request bypasses the meter. The power sampler reads the device's own telemetry, at a rate high enough to resolve a prefill burst of one or two seconds. It measures the inference device rather than the agent host, so the energy it integrates is that of the model and not of the simulator. The verification host holds the independent verification script, the fixed acceptance criteria and the record of runs, none of which the agent can see. A cloud model replaces the inference node behind the same executor, tools and verification script, and then reports tokens and time only. The context window, the output-token budget and the sampling rate are stated with every batch, since each enters the four quantities. Section~\ref{sec:cost} lists how each role is filled in this study.

\begin{figure}[!t]
\centering
\includegraphics{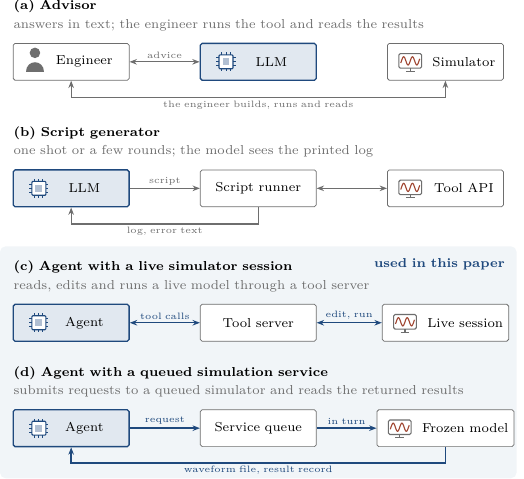}
\caption{Four ways a language model is coupled to engineering software. In (a) and (b) the engineer or a fixed script runs the iteration; (c) and (d) close the engineering loop and are the two couplings used in this paper.}
\label{fig:interaction}
\end{figure}

As shown in Fig.~\ref{fig:interaction}, a language model can be coupled to engineering software in four ways, and the coupling determines what the model can observe and how its result can be checked. As an advisor, the model answers in text and the engineer operates the tool. As a script generator, it writes a script or netlist that a runner executes through the tool's programming interface, and it sees the printed log. In both, the engineer or a fixed script runs the iteration, while the other two close the loop of Fig.~\ref{fig:workflow} and are the two used here. Both reach the simulator through a tool protocol, in which a server provides named tools with typed arguments, and the executor places those descriptions in the model's context. However, they differ in whether the agent edits a live model or submits requests against a frozen one. In a live-session coupling, the tool server attaches to an open simulator session, one per agent, and provides tools that read a model, edit it, run code and check the result. A time limit on each tool call bounds a simulation that does not return, and tool calls are approved automatically so that the loop runs unattended. The model never sees a block diagram. Instead, it reads the model as text returned by a tool, and edits are sent in the opposite direction as structured arguments or code. A simulation returns whatever the tool or the agent's own script computes from the logged signals. The quality of the feedback the agent receives is therefore limited by the tools, and because every tool result is re-sent on every later turn, a verbose tool output adds prefill work for the remainder of the run. In a simulation-service coupling, the agent never opens the model. A service on the simulator's host holds the licence, queues the requests of all agents, runs them in turn and returns the requested output channels as a file with a result record. The agent host carries only the service client. The deliverable is then a setting, a selection, a rule or a number rather than a model file, and the time a request waits in the queue is part of the cost. The first coupling suits a simulator that offers a scripting session to a tool server, while the second suits one whose licence or host constrains where it runs.

\subsection{Task Sets}

\begin{figure*}[!t]
\centering
\includegraphics{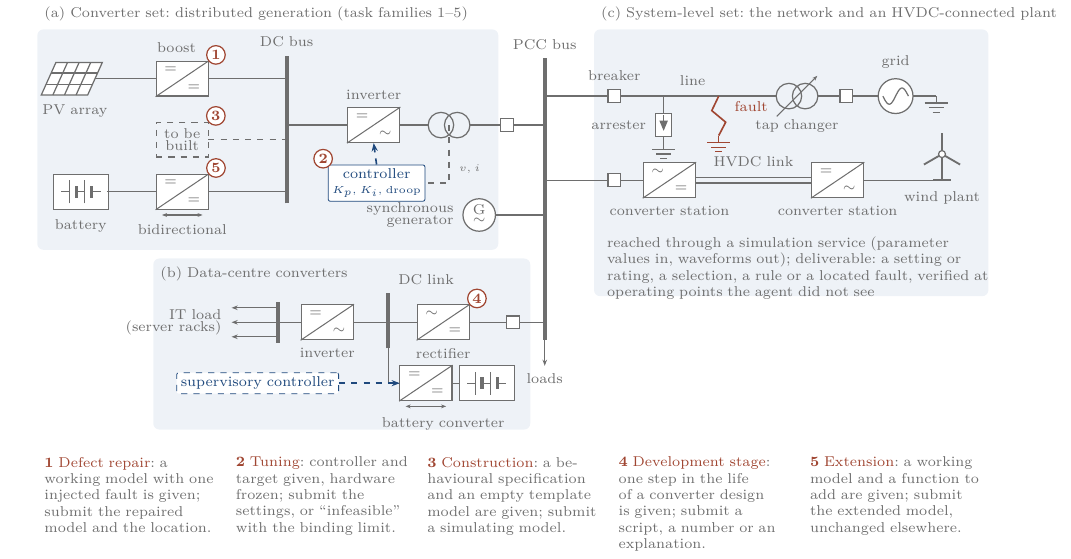}
\caption{The task sets drawn on one converter-interfaced microgrid, in single-line form. (a) Converter set: distributed generation behind a DC bus and a grid-forming inverter; the circled numbers mark where the five task families act, defined below the diagram. (b) Data-centre converter system: the chain on which the development-stage family and the data-centre set work. (c) System-level set: the network the microgrid connects to, with the line, protection, transformer and HVDC-connected plant that the studies concern; the agent never opens these models and reaches them through a simulation service. The diagram is generic and reproduces no task.}
\label{fig:taskcards}
\end{figure*}

A task for this procedure needs a written specification, a deliverable that can be simulated, and a numerical check at operating points the agent does not see, and any simulator and task domain that allow these can be used. Two task sets are used here. As summarized in Table~\ref{tab:tasks}, the converter set has five families of tasks on MATLAB/Simulink models, each with a written behavioural specification that the verification script checks numerically. The five are routine tasks of a converter engineer. Defect repair asks for a faulty model to be diagnosed and corrected. Controller tuning asks for controller settings with the hardware fixed, either in a single-shot form or in a multi-round form in which the agent sees the simulated result after each proposal. Where a specification cannot be met, the correct deliverable says so. Construction asks for a model to be built from a specification, starting from an empty template model. Development-stage tasks are steps from the lifecycle of a data-centre converter design, and extension tasks add a function to a working model. The models range from a few hundred to about three thousand blocks. A further set of six data-centre converter tasks was run on the cloud model alone, to compare its reasoning tiers, and it is counted in the dataset total but in none of the local test conditions. Fig.~\ref{fig:taskcards} illustrates what each family gives the agent and what it must submit, and the form in which the system-level tasks of the second set reach it.

\begin{table}[!t]
\caption{Converter Set: Deliverable and Verified Aspects per Family (All Must Pass)}
\label{tab:tasks}
\centering
\footnotesize
\renewcommand{\arraystretch}{1.12}
\begin{tabular}{@{}>{\raggedright\arraybackslash}p{1.75cm}>{\raggedright\arraybackslash}p{2.45cm}>{\raggedright\arraybackslash}p{3.4cm}@{}}
\toprule
Family & Deliverable & Verified on \\
\midrule
Defect repair & Repaired model and a diagnosis & Diagnosis; behaviour; minimality of the change \\
Controller tuning & Settings, or a statement that the specification cannot be met & Behaviour; interface; footprint; stated limit \\
Construction & A simulating model & Behaviour; interface; structure \\
Development stage & Script or attribution, with simulation & Agreement with a reference analysis \\
Extension & Extended model & Behaviour; interface; footprint \\
\bottomrule
\end{tabular}
\end{table}

The second set extends the scope from the converter to the systems it connects to. It has thirteen tasks in PSCAD/EMTDC, which are listed with their identifiers in Table~\ref{tab:pscad} (Appendix~\ref{app:pscad}). Six concern converter-interfaced resources and HVDC, and the other seven concern the network studies against which such installations are assessed, from clearing times and stabilizer tuning to switching transients and insulation coordination. The deliverables are of the kind a planning, protection or compliance engineer produces. The reference models are published examples (the simulator vendor's knowledge base and cookbook, the PNNL i2x library, an open grid-forming inverter library and the CIGRE HVDC benchmark). Each deliverable is re-simulated or compared with a reference result at operating conditions the agent does not see.

\subsection{Procedure and Fairness Safeguards}
\label{sec:fairness}

The Algorithm lists the test procedure. Before a batch, the task set, the acceptance criteria and the undisclosed operating points are frozen; the output-token budget, the tool set and the executor version are fixed; and the test conditions to be run are listed. Each run then follows seven steps: 1) a link self-check; 2) a wait until the inference device has returned to its idle power, so that the energy integral starts from a known state; 3) clearing of the workspace, with only the task's own files copied in; 4) the run, during which the agent sees the task text, the tools and the budget and nothing else; 5) collection of the executor's session log, the metering proxy's request record, the power trace and the final workspace; 6) verification by the independent script; and 7) recording of the outcome with the four quantities.

\begin{algorithm}[!t]
\caption{Test procedure for one batch}
\label{alg:procedure}
\footnotesize
\begin{algorithmic}[0]
\Statex \textbf{Before the batch}
\State freeze the task set, the acceptance criteria and the
\Statex \hskip 1.6em undisclosed operating points
\State fix the output-token budget, the tool set and the executor version
\State list the test conditions: task $\times$ model $\times$ configuration $\times$ repetitions
\Statex
\Statex \textbf{Every run}
\ForAll{test condition and repetition}
  \Repeat
    \State \textbf{1:} link self-check: tool round trip and model health
    \State \textbf{2:} wait until the inference device is back at its idle power
    \State \textbf{3:} wipe the workspace; copy in only the files of this task
    \State \textbf{4:} run the agent under the token budget
    \AlgNote{the agent sees the task, the tools and the budget only}
    \State \textbf{5:} collect the session log, the request record, the power
    \Statex \hskip 5.4em trace and the final workspace
    \State \textbf{6:} verify the deliverable with the independent script
    \AlgNote{the agent's own report is ignored}
  \Until{the run is valid, or it has been re-run three times}
  \State \textbf{7:} record the outcome with power, energy, tokens and time
  \AlgNote{pass, fail, timeout, no final answer or invalid;}
  \AlgNote{an invalid run (infrastructure fault) is never counted}
\EndFor
\end{algorithmic}
\end{algorithm}

The following rules make a comparison between two models, or between a local and a cloud deployment, fair, and the results of Section~\ref{sec:cost} rest on them. Every model sees the same task text, the same tools and the same limits. Criteria, tolerances, undisclosed cases and the rules for reading a result are written before the batch and not revised after a result is seen. When an agent finds a defect in a task, the task is revised and re-run, and the attempt that found it is not re-judged. Where a reference solution exists, the verification script is checked before it is frozen against reference submissions whose outcome is known in advance. The agent's own report of success is not accepted as evidence. Instead, the verification script recomputes the result, where possible at operating points the agent did not see. Undisclosed operating points and verification independent of the agent's report are also the basis of PowerAgentBench \cite{R20_poweragentbench_ss,R21_poweragentbench_dyn}. However, its separate scores for the evidence behind a submission, and for a submission made after the agent itself had measured a failing case, are not adopted here. Submissions of the latter kind are recorded with the power system results (Appendix~\ref{app:pscad}). An invalid run is one stopped by an infrastructure or author-side fault, and invalidated attempts are re-run in the same test condition, at most three times, rather than discarded. A disturbance that can only make the task harder, e.g., a slower shared queue, leaves a pass valid and invalidates a failure. Isolation is by construction, and in the power system set every tool output is checked for leakage of answer material. Any detected leakage invalidates the attempt regardless of its outcome. None of these rules is specific to the model under test.

\subsection{Verification, Records and Dataset}

Each run ends as a scored pass, a scored fail, a timeout, a run with no final answer, or an invalid run. The success rate $p$ and the test condition are as defined in Table~\ref{tab:terms}. Differences in pass counts between two test conditions are tested with a two-sided exact test stratified by task, which is Fisher's exact test when the test conditions hold one task. Its value is written $P$ to distinguish it from the success rate $p$. Intervals on success rates are Wilson 95\% intervals. Confidence intervals on energy per solved task are percentile bootstrap intervals over the runs of a test condition (2,000 resamples). For a test condition with one or two passes, the interval is conditional on at least one pass and understates the uncertainty. Power is the power of the inference device, excluding the agent host, and energy is its integral over the run window, with verification time excluded. Token counts are taken from the executor's session log. Reasoning tokens and cache hits are counted separately. Energy per solved task is defined per test condition as total energy divided by the number of passes. The power system set has one cloud attempt and two local attempts per task.

\section{Measured Capability and Costs}
\label{sec:cost}

\begin{figure*}[!t]
\centering
\includegraphics{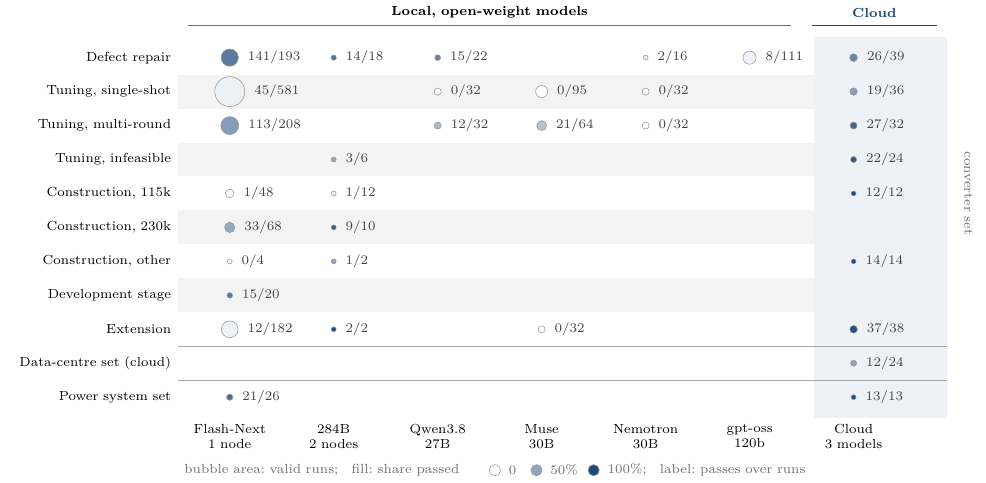}
\caption{The campaign by test condition. Bubble area is proportional to the number of valid evaluated runs, fill to the pass fraction, and the label gives passes over runs; the converter set holds 2,017 runs and the data-centre set 24 cloud runs, 2,041 in all, and the power system set 26 local and 13 cloud attempts.}
\label{fig:coverage}
\end{figure*}

In this section, the implications of Section~\ref{sec:mechanism} are tested against the measurements, as summarized in Table~\ref{tab:implications}. The capability boundary is presented first, because the success rate $p$ scales every other cost through \eqref{eq:esuccess}. Then, the four quantities are presented in turn, followed by the cost per solved task, the comparison with the cloud model and the interventions on the loop. The runs behind the results are counted in Table~\ref{tab:dataset}, and Table~\ref{tab:instance} lists how the roles of Section~\ref{sec:attachment} are filled. Fig.~\ref{fig:coverage} shows the campaign by test condition.

\begin{table}[!t]
\caption{The Implications of Section~\ref{sec:mechanism} and the Corresponding Measurements}
\label{tab:implications}
\centering\footnotesize
\setlength{\tabcolsep}{3pt}\renewcommand{\arraystretch}{1.1}
\begin{tabular}{@{}r>{\raggedright\arraybackslash}p{2.55cm}>{\raggedright\arraybackslash}p{3.45cm}l@{}}
\toprule
 & Implication & Measurement & Sec. \\
\midrule
1 & Decode power is a plateau that the task does not alter & 33--38\,W across families, budgets and outcomes & \ref{sec:power} \\
2 & Prefill is higher in power and short & About 1.4\,pu; 8.8\% of busy time & \ref{sec:power}, \ref{sec:energy} \\
3 & Nominal input tokens overstate the computation & 6.0\% of nominal tokens computed & \ref{sec:tokens} \\
4 & Reasoning is decoded and metered as output & About three quarters of local output; no lower power with reasoning disabled & \ref{sec:power}, \ref{sec:tokens} \\
5 & Energy follows duration, then output tokens & Median error 2.4\% from duration alone, 38 held-out runs & \ref{sec:energy} \\
6 & The model does not see its token consumption & With a notice, cap stops fell from 10 of 24 to 2 of 16 & \ref{sec:tokens} \\
7 & Time is model time plus tool time & Device busy 94\% of the window & \ref{sec:timeres} \\
\bottomrule
\end{tabular}
\end{table}

\begin{table}[!t]
\caption{Summary of the Runs Underlying the Results}
\label{tab:dataset}
\centering\footnotesize
\begin{tabular}{@{}lr@{}}
\toprule
Converter and data-centre sets, valid evaluated runs & 2,041 \\
\quad converter set & 2,017 \\
\quad data-centre set, cloud only & 24 \\
\quad passes & 617 \\
\quad two earlier cloud models, pooled in the cloud column & 183 \\
\quad with an energy record (70,392\,kJ, 19.6\,kWh) & 882 \\
\quad no energy record: Flash-Next tuning & 789 \\
\quad no energy record: cloud & 219 \\
\quad no energy record: local, sampler not running & 151 \\
Power system set, local attempts & 26 \\
Power system set, cloud attempts & 13 \\
\bottomrule
\end{tabular}
\end{table}

\begin{table}[!t]
\caption{The Framework as Instantiated in This Study}
\label{tab:instance}
\centering\footnotesize
\renewcommand{\arraystretch}{1.1}
\begin{tabular}{@{}>{\raggedright\arraybackslash}p{1.75cm}>{\raggedright\arraybackslash}p{6.25cm}@{}}
\toprule
Role & Instance \\
\midrule
Executor & Codex command-line agent \cite{R70_codex}, pinned per deployment: 0.153.3 local, 0.159.3 cloud \\
Live-session coupling & MATLAB/Simulink through the vendor's agent toolkit: thirteen MCP tools (five read a model, one edits it, three run code, four check), one MATLAB session per agent, 600\,s per tool call \\
Service coupling & PSCAD/EMTDC service on a Windows host with one licence, run one request at a time; 10 to 170\,s per simulation \\
Inference node & NVIDIA DGX Spark: GB10 single-package device, 140\,W thermal design power, 128\,GB unified memory shared by weights and context, 273\,GB/s; vLLM \cite{R71_vllm}; two nodes with tensor parallelism over a direct 200\,Gb/s link for the larger model \\
Main local model & Flash-Next \cite{R69_flashnext}, mixture of experts in NVFP4: 176B total, 6B active per token, about 126\,GB of weights; a second NVFP4 conversion of 135\,GB for the power system set \\
Other local models & DeepSeek V4 Flash \cite{R73_dsv4flash}, 284B (13B active), on two nodes; Qwen3.8-27B \cite{R74_qwen27b} and Muse Glimmer 30B \cite{R75_muse} (dense); Nemotron 3.5 Lightning 30B \cite{R76_nemotron}; gpt-oss-120b \cite{R77_gptoss} \\
Cloud models & gpt-6.1-sol \cite{R78_gpt61sol}; two earlier cloud models, pooled with it in the cloud column of Figs.~\ref{fig:coverage} and~\ref{fig:ladder} \\
Context, sampling & 262,144-token context window; device telemetry at 1 or 10\,Hz, by batch \\
\bottomrule
\end{tabular}
\end{table}

\subsection{Capability Boundary}
\label{sec:capability}

\begin{figure}[!t]
\centering
\input{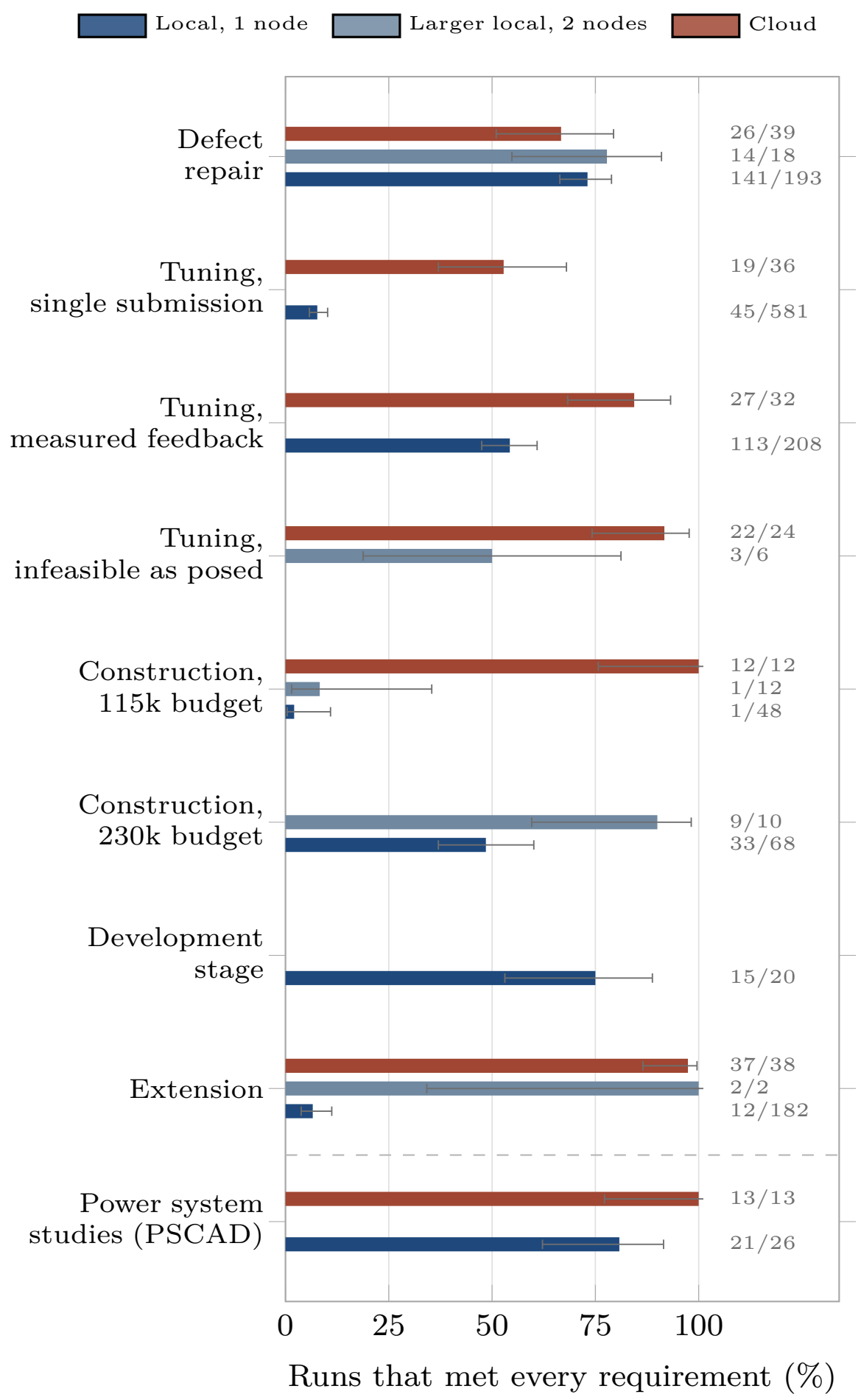}
\caption{Where capability ends: the share of valid evaluated runs that met every requirement under independent verification, by task family and deployment, over both task sets; labels give passes over runs and whiskers are Wilson 95\% intervals. The power system row pools the two local attempts per task.}
\label{fig:ladder}
\end{figure}

The main result on capability is that the local agent solves most tasks that start from an existing model, and few that start from a specification. Fig.~\ref{fig:ladder} gives the success rate of every task family in both sets, for the main local model on one node, the larger model on two nodes and the cloud models. Where the agent diagnoses and repairs an existing model or carries out a step of an existing design, the local model passes most runs, i.e., 73\% of defect repairs, including bases of three thousand blocks, and 75\% of development-stage tasks. In contrast, where it must produce a design from a specification, it mostly fails under the stated conditions: 8\% of single-submission tuning, 2\% of construction at the 115k budget and 7\% of extensions. Two changes to the loop have the largest effect on these families, i.e., measured feedback (tuning from 8 to 54\%) and a larger budget (construction from 2 to 49\%, and 90\% on two nodes). Sections~\ref{sec:tokens} and~\ref{sec:tradeoffs} return to them. In the converter set, the cloud models pass 84 to 100\% of the families they were run on, except defect repair (67\%, run on an earlier cloud model only) and single-submission tuning (53\%). The power system studies, whose deliverable is a setting or a number found by simulating a given model, behave like the repair and development families rather than like construction.

For the power system set, the cloud reference is gpt-6.1-sol at its lowest reasoning tier, with one attempt per task, a 3\,h limit and no token budget. Its role is to establish that each task can be solved through the interface. The local column is Flash-Next on the single node with the same limit and two attempts per task. Every attempt is listed in Section~\ref{sec:timeres}.\label{sec:pscad}

The cloud reference passed all thirteen tasks. The local model passed nine on the first attempt and twelve on the second, and wherever both deployments passed, they reached the same result. The local failures did not arise from the simulation work itself. One first attempt reached the time limit without a deliverable. The other three were submitted with the failing evidence inside the report, and the model did not act on that evidence before submitting, by revising the setting, testing a longer duration or stating that the requirement could not be met. In a supervised workflow, that check is made by the engineer who reads the report. The attempts, the failures and the defects in the tasks that the batches exposed are described in Appendix~\ref{app:pscad}, which also shows waveforms from the runs (Fig.~\ref{fig:app_waves}).

\subsection{Power}
\label{sec:power}

As an electrical load, the inference device has three levels: an idle power, a constant decode power and a short prefill peak. By Implication~1, the device should draw the same power while decoding regardless of the agent's activity, as long as the serving settings are left unchanged, and the measurements confirm this. Power here is the device's own telemetry for its GPU domain, which excludes the rest of the system. Therefore, the figures are those of the inference computation and lie below the power at the AC input. Across twelve batches and more than three hundred measurements in the first month, the single-node device drew 33--38\,W while decoding, i.e., about a quarter of the thermal design power of its package (Table~\ref{tab:instance}), against an idle power of 13--14\,W. In these measurements, the plateau did not vary with the task family, the token budget, the number of undisclosed operating points or whether the run reached its cap. Power is given below in watts and, where models or devices are compared, in per unit of the model's own decode plateau, $P_\mathrm{dec} = 1$\,pu. On this base, Flash-Next's idle power is about 0.37\,pu and its prefill level about 1.4\,pu, and the mean power of a run is approximately $0.37 + 0.63\,d$\,pu for a share $d$ of the window spent generating. Thus, a record from another model or device can be compared on the same base. Each node of the two-node configuration drew 31--37\,W, and disabling reasoning did not lower the power. As shown in Table~\ref{tab:strata}, the two dense models, which read more weight per token, have the highest window-mean power. It should be noted that quantization and serving settings were not controlled across models, so this ordering is consistent with the mechanism rather than a test of it.

\begin{figure}[!t]
\centering
\input{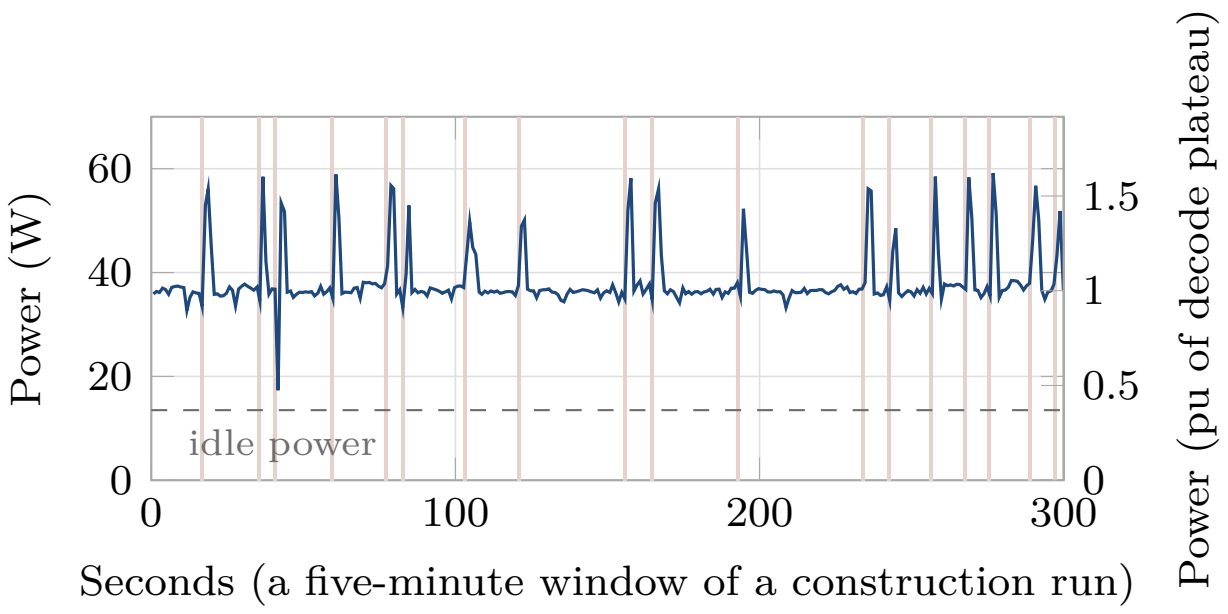}
\caption{Five minutes of a passing construction run (Flash-Next, single node, 230k budget), device power at 1\,Hz. Shaded strips are the agent's tool calls; each is followed by a prefill spike of 50--59\,W, and the model decodes at the 36\,W plateau until its next call.}
\label{fig:zoom}
\end{figure}

\begin{figure}[!t]
\centering
\input{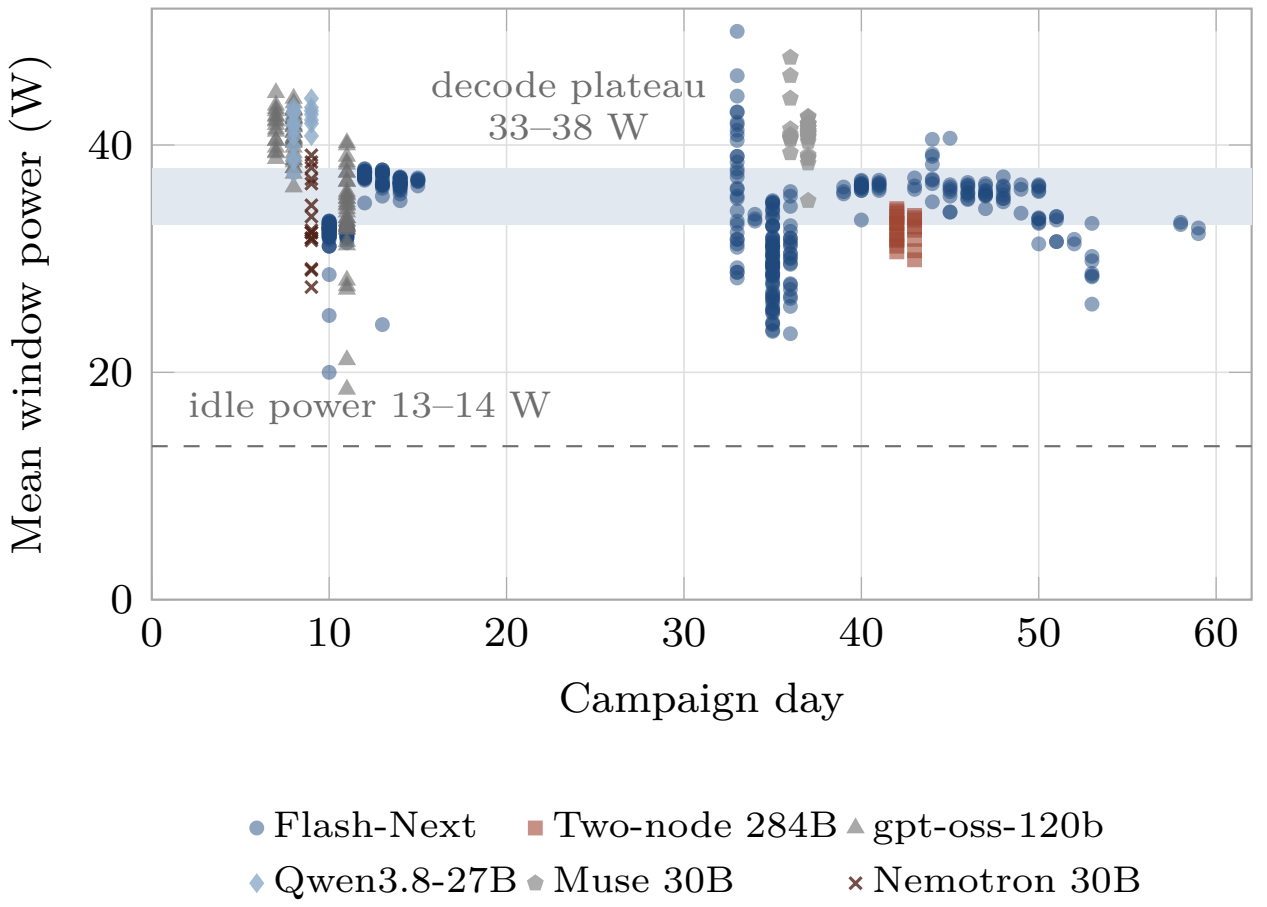}
\caption{Mean GPU power over the run window for every run with a window-mean power record on the main inference device in the defect, construction, development-stage and extension families ($n = 566$), by model and campaign day. Window means lie below the decode plateau of the respective model by the share of tool-waiting time; the band marks the Flash-Next plateau.}
\label{fig:power}
\end{figure}

Prefill is compute-bound. In a controlled sweep of input lengths, it drew 1.59 times the decode power above idle, and its power depended on die temperature, at 0.601\,W/$^\circ$C, and not on input length. As shown in Fig.~\ref{fig:zoom}, within a run it appears as a spike of one or two seconds each time a tool returns and the model reads the new context, with the decode plateau between. Temperature is also the reason the plateau is a band rather than a single value. Flash-Next's decode power above idle drifted by about 25\% across nine batches. Fig.~\ref{fig:power} shows the consequence of a constant plateau for the four families whose runs consist mostly of model time. Their window means lie a few watts below the plateau, by the share of time spent waiting for tools. In the multi-round tuning test conditions, where the verification simulation ran inside the window, the means fall to 13.6 and 16.9\,W. These lower means are caused by lower duty cycles, not by lower plateaus. Therefore, a load model of an engineering agent has to represent the plateau and the duty cycle separately.

\begin{table}[!t]
\caption{Flash-Next Defect Runs by Base-Model Scale and Window-Mean Power by Model in the Four Model-Time Families}
\label{tab:strata}
\centering
\footnotesize
\setlength{\tabcolsep}{2.6pt}
\begin{tabular}{@{}lrrrrr@{}}
\toprule
Base-model scale & Blocks & Runs & Pass & \begin{tabular}[b]{@{}r@{}}Median\\kJ/run\end{tabular} & \begin{tabular}[b]{@{}r@{}}kJ per\\solved task\end{tabular} \\
\midrule
Bridges & $\sim$10$^2$ & 98 & 81 & 30.2 & 46.0 \\
PV boost, relay & 563/753 & 48 & 33 & 95.5 & 139.0 \\
EV grid-tied node & 3,025 & 47 & 27 & 105.8 & 191.9 \\
\midrule
Model & Runs & \begin{tabular}[b]{@{}r@{}}p10\\(W)\end{tabular} & \begin{tabular}[b]{@{}r@{}}Median\\(W)\end{tabular} & \begin{tabular}[b]{@{}r@{}}p90\\(W)\end{tabular} & \\
\midrule
Flash-Next & 384 & 28.6 & 33.7 & 37.5 & \\
gpt-oss-120b & 90 & 32.8 & 39.3 & 43.0 & \\
Qwen3.8-27B & 22 & 38.9 & 42.2 & 43.3 & \\
Muse 30B & 32 & 38.9 & 41.0 & 42.5 & \\
Two-node 284B$^{a}$ & 22 & 30.7 & 32.8 & 34.0 & \\
Nemotron 30B & 16 & 29.1 & 32.4 & 38.2 & \\
\bottomrule
\end{tabular}
\par\smallskip\raggedright\scriptsize $^{a}$Power per node.
\end{table}

\subsection{Energy}
\label{sec:energy}

\begin{figure}[!t]
\centering
\input{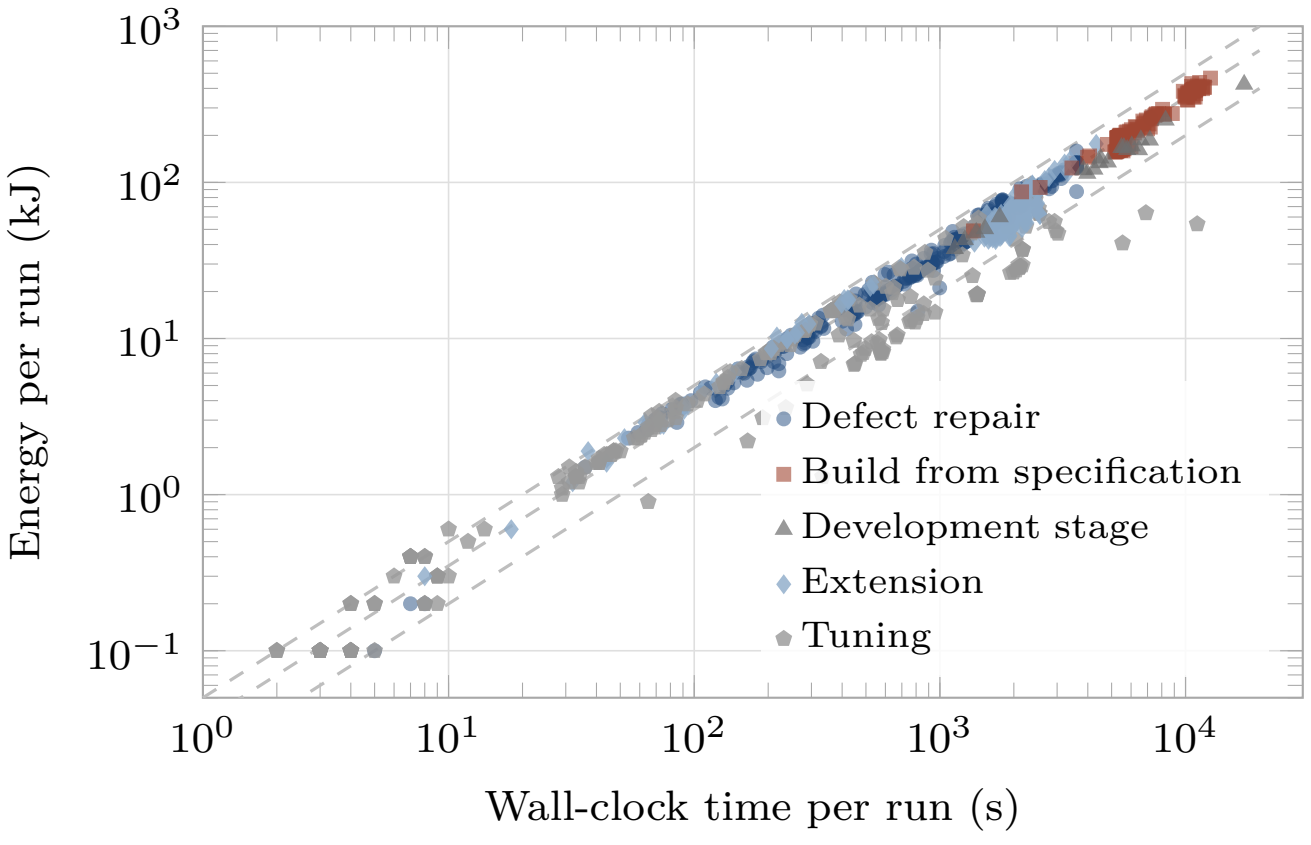}
\caption{Energy versus wall-clock time per run on the single-node device ($n = 832$). Dashed lines are constant-power references at 20, 35 and 50\,W (bottom to top). Tuning runs fall below 20\,W because verification wait time dominates their windows.}
\label{fig:energytime}
\end{figure}

\begin{figure}[!t]
\centering
\input{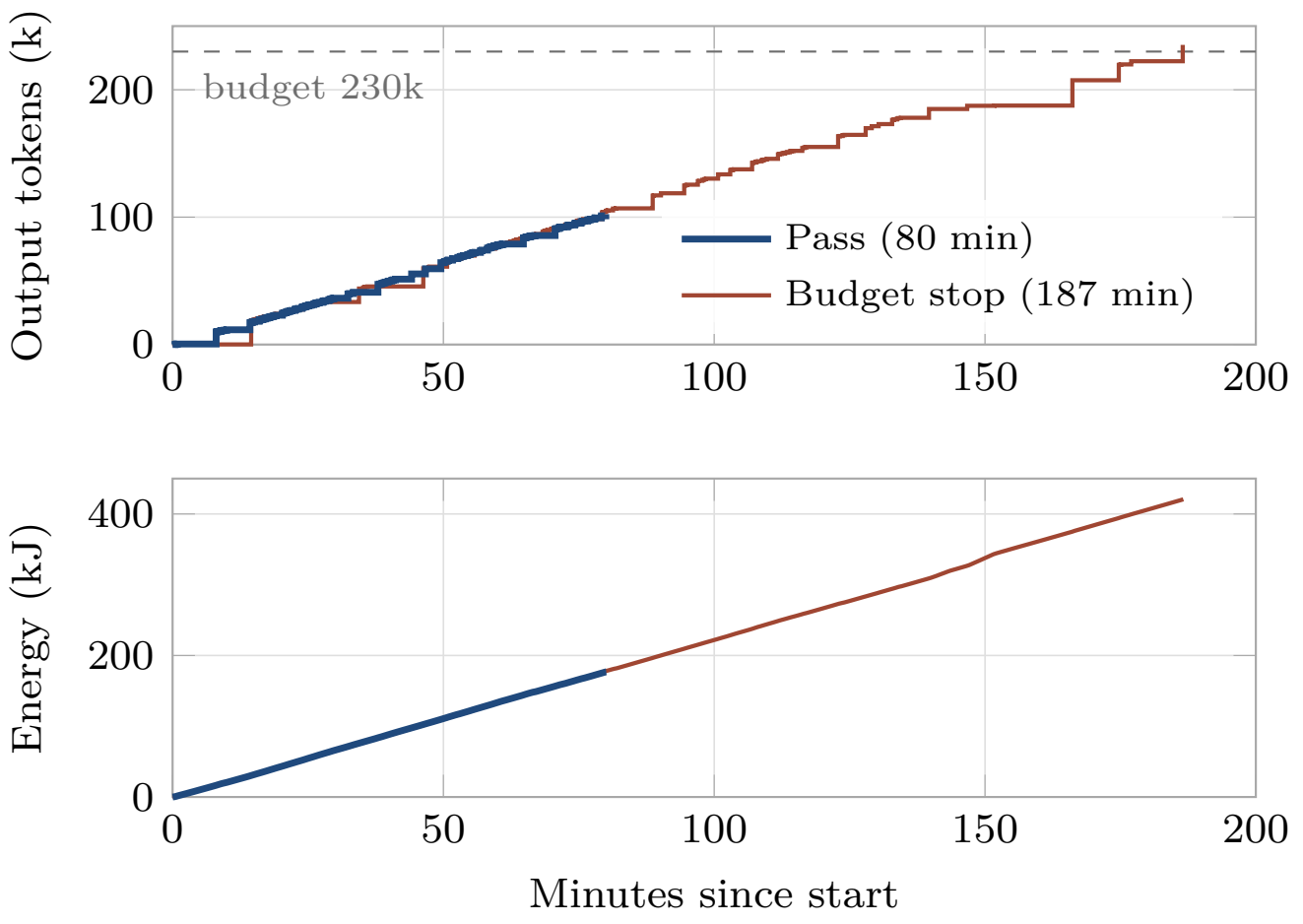}
\caption{A pass and a budget stop on the same device, model and 230k budget, from one batch: cumulative output tokens (top) and cumulative energy from 1\,Hz samples (bottom). The failing run decodes at the same rate until it is stopped at the cap, and consumes about 2.4 times the energy of the passing run.}
\label{fig:capstop}
\end{figure}

Because the plateau is flat, the energy of a run is set almost entirely by its duration. This is confirmed in Fig.~\ref{fig:energytime}. For every single-node run with both records, energy plotted against wall-clock time falls on a line whose slope is the plateau. The median energy per run rises with the time each family needs, from 62.9\,kJ for a defect run to 353.9\,kJ for a construction run at the 230k budget. The token budget sets an upper bound on duration. For example, a construction run that uses its whole 230k budget costs about 400\,kJ with reasoning enabled, whereas the runs that passed, which stopped before the cap, cost 175--351\,kJ. Across the fourteen test conditions with both outcomes and an energy record, a failing run cost 0.94 to 2.9 times a passing run on average. This supports the assumption behind \eqref{eq:esuccess} that a failed run costs about as much as a passing run or more. Fig.~\ref{fig:capstop} shows one pair from a single batch, i.e., a construction run that passed after 80\,min and 102k output tokens, and a run on a harder task that decoded at the same rate until it had spent the 230k budget, at 187\,min. Equation~\eqref{eq:energy} was tested on 38 held-out runs from five batches by reconstructing the energy of each run from its duration alone, from the token stream with fitted phase constants, and from the measured request timestamps. The median errors were 2.39\%, 1.89\% and 2.63\%, respectively, which supports the decomposition, and the duration alone suffices. In the same batches, prefill occupied 8.8\% of the device's busy time, and busy time was 94\% of the window, which is the direct test of Implication~2. Therefore, estimating the energy of a group of agents needs only their durations. As a load, a run reduces to 1) an idle power; 2) a decode plateau with a temperature-dependent band; 3) a prefill level of about 1.4\,pu that lasts a small fraction of the time; and 4) a duty cycle set by the alternation of model and simulator. Thus, the power spectrum of an engineering agent is determined by the timing of the simulator rather than of the model.

On the power system set, the device power trace, sampled at 1\,Hz for the twelve second local attempts it covers, gives 42 to 102\,Wh per attempt (median 79) at a window mean of 25 to 34\,W. In these attempts, the device is at its idle power for 20\% of the time, which is the time spent waiting for the simulator. Note that these are the figures for one small inference device, i.e., the smallest configuration on which the study could be carried out, not for a deployment sized for throughput (Section~\ref{sec:wherecost}).

\begin{table}[!t]
\caption{Four Records of Two Passing Local Runs and of the Cloud Reference (Lowest Reasoning Tier) on the Same Tasks; G1 Simulations Are the 31 \texttt{run\_matlab\_file} Calls}
\label{tab:breakdown}
\centering
\scriptsize
\setlength{\tabcolsep}{3pt}
\renewcommand{\arraystretch}{1.05}
\begin{tabular}{@{}p{2.95cm} rr rr@{}}
\toprule
 & \multicolumn{2}{c}{G1 construction} & \multicolumn{2}{c}{S12 fault location} \\
\cmidrule(lr){2-3}\cmidrule(lr){4-5}
 & Local & Cloud & Local & Cloud \\
\midrule
\multicolumn{5}{@{}l}{\textit{Time}} \\
Duration (min) & 80 & 3 & 177 & 11 \\
Model time, transcript (min) & 77 & 2 & 152 & 3 \\
Tool time (min) & 2.4 & 0.5 & 25.3 & 7.5 \\
Busy / idle, power trace (min) & 78 / 2 & -- & 151 / 26 & -- \\
Simulations (count, total min) & 31, 1.5 & -- & 24, 64 & -- \\
\multicolumn{5}{@{}l}{\textit{Tokens}} \\
Requests / tool calls & 153 / 152 & 16 / 15 & 76 / 75 & 59 / 58 \\
Input, nominal (M) & 13.95 & 0.44 & 7.91 & 1.33 \\
\quad served from cache (\%) & 96.6 & 91.8 & 98.9 & 96.0 \\
\quad computed (k) & 478 & 36 & 90 & 53 \\
Output tokens (k) & 102.3 & 6.2 & 142.3 & 7.2 \\
\quad reasoning share (\%) & 74 & 12 & 81 & 19 \\
Peak context (k) & 165 & 34 & 204 & 30 \\
Output tokens / model s & 22.1 & 55.2 & 15.6 & 38.2 \\
\multicolumn{5}{@{}l}{\textit{Power and energy}} \\
Idle power / plateau (W) & 15.3 / 36.1 & -- & 13.3 / 34.2 & -- \\
Mean power (W) & 36.4 & -- & 31.1 & -- \\
Energy (kJ) & 175 & -- & 331 & -- \\
\quad idle term $P_\mathrm{idle}T$ (kJ) & 73 & -- & 141 & -- \\
\quad above idle (kJ) & 102 & -- & 190 & -- \\
Energy while idle (kJ) & 2 & -- & 21 & -- \\
Energy per output token (J) & 1.71 & -- & 2.33 & -- \\
\quad above idle (J) & 0.99 & -- & 1.33 & -- \\
\bottomrule
\end{tabular}

\end{table}

\begin{figure}[!t]
\centering
\input{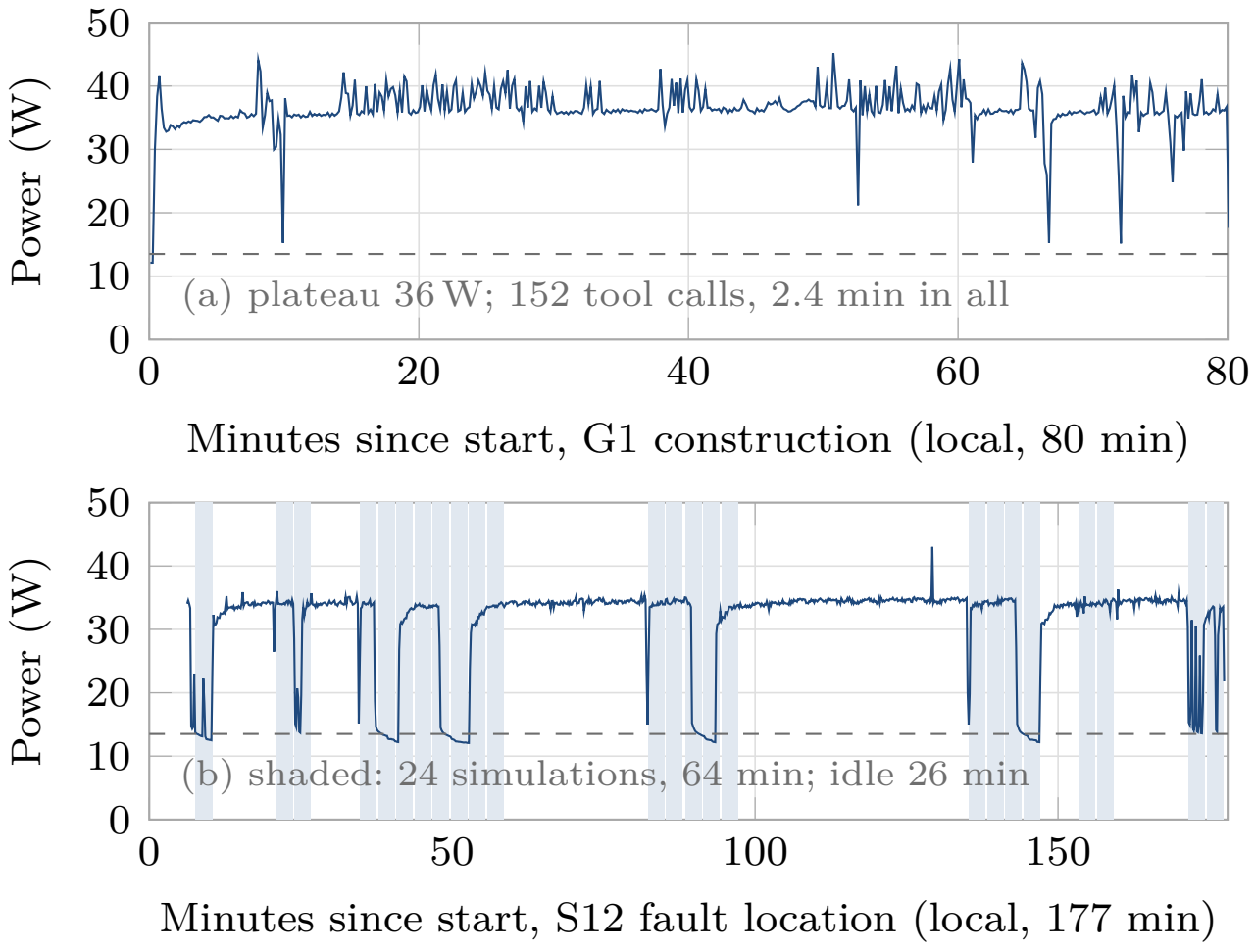}
\caption{Device power (10\,s means) over the window of the two local runs of Table~\ref{tab:breakdown}. (a) Converter construction G1: the device stays at the decode plateau because tool calls return in seconds. (b) Fault location S12: shaded intervals are the 24 PSCAD simulations, and the sampler started 6.4\,min into the window; the device falls to the idle power only during the agent's blocking waits, since simulations ran in the background while it continued to generate.}
\label{fig:breakdown}
\end{figure}

Table~\ref{tab:breakdown} and Fig.~\ref{fig:breakdown} follow two passing local runs through all four quantities, with the cloud reference on the same task beside each. The two runs are a converter construction run (G1, Flash-Next at the 230k budget, 80\,min) in the Simulink coupling, and the second local attempt on the fault-location task (S12, 177\,min) in the service coupling. Time is determined in two ways, i.e., from the transcript as model time plus tool time, and from the power trace as the seconds above and below the idle power. The two agree within a minute on both runs. The converter run kept the device busy for 78 of 80\,min, because its 152 tool calls, 31 of them simulations, took 2.4\,min in all. The power-system run was idle for 26 of 177\,min although its 24 simulations occupied 64\,min. This is because the agent submitted each simulation to the service in the background and continued reasoning, polling and writing while it ran. As shown in Fig.~\ref{fig:breakdown}(b), only its blocking waits of 4 to 5\,min each appear as idle time. Therefore, the idle share in Implication~5 is set by how the executor waits for the simulator, not by the simulation time that the task requires.

Three quarters to four fifths of the local output was reasoning. The cloud reference produced about one twentieth of the output tokens on the same tasks, with a reasoning share of 12 and 19\%, and on S12 spent more of its wall-clock time waiting for the simulator than generating. The energy decomposes as \eqref{eq:energy}. On the converter run, the idle term $P_\mathrm{idle}T$ is 73\,kJ of 175, and the 102\,kJ above it is almost entirely decode, which gives 1.0\,J per output token above idle. On the power-system run, the idle term is 141 of 331\,kJ and the idle seconds during blocking waits cost 21\,kJ at idle power. Thus, the marginal energy of an output token is about 1\,J on this device, and the remainder of the energy is the idle power multiplied by the run duration. The two runs differ in energy by 1.9 times and in duration by 2.2 times, the longer run having spent more of its window at idle power. One difference in conditions must be noted with the comparison. The power-system set was served from a second NVFP4 conversion of the weights and a server build rebuilt with a 262k context, and it decoded about 30\% more slowly than the converter set. The cause was not isolated.

\subsection{Tokens and the Token Budget}
\label{sec:tokens}

\begin{figure}[!t]
\centering
\input{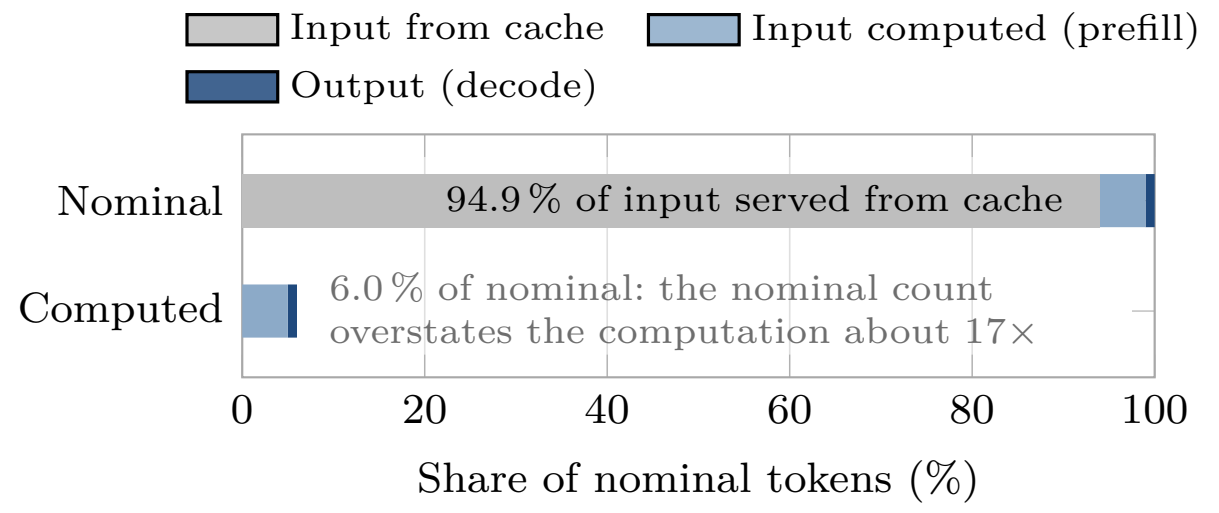}
\caption{The token record of 693 runs from one month of the campaign. The executor counts every input token on every request, but 94.9\% of the input was served from cache; the tokens actually computed, prefill of the uncached input plus decode of the output, are 6.0\% of the nominal total.}
\label{fig:tokens}
\end{figure}

Tokens are the one quantity recorded by the agent software itself. As Implication~3 states, the nominal count overstates the computation. As shown in Fig.~\ref{fig:tokens}, over 693 runs from one month of the campaign, 99.0\% of the nominal tokens were input, and 94.9\% of that input was served from cache, so only 6.0\% of the nominal total was actually computed. Therefore, an energy estimate from nominal totals would be too high by a factor of about seventeen. Within one passing construction run of 153 requests, the input counted on each request is the whole context, rising to 164k tokens. In contrast, the part actually computed in this run is what the last tool returned, a median of 2.9k tokens, and a request's output is a few hundred tokens. About three quarters of the local output is reasoning, compared with about half for the cloud model on the same construction tasks. On the two-node model, single requests reached about 244k tokens, against a usable window of 258k in the executor (the server context was 262,144).

\begin{figure}[!t]
\centering
\input{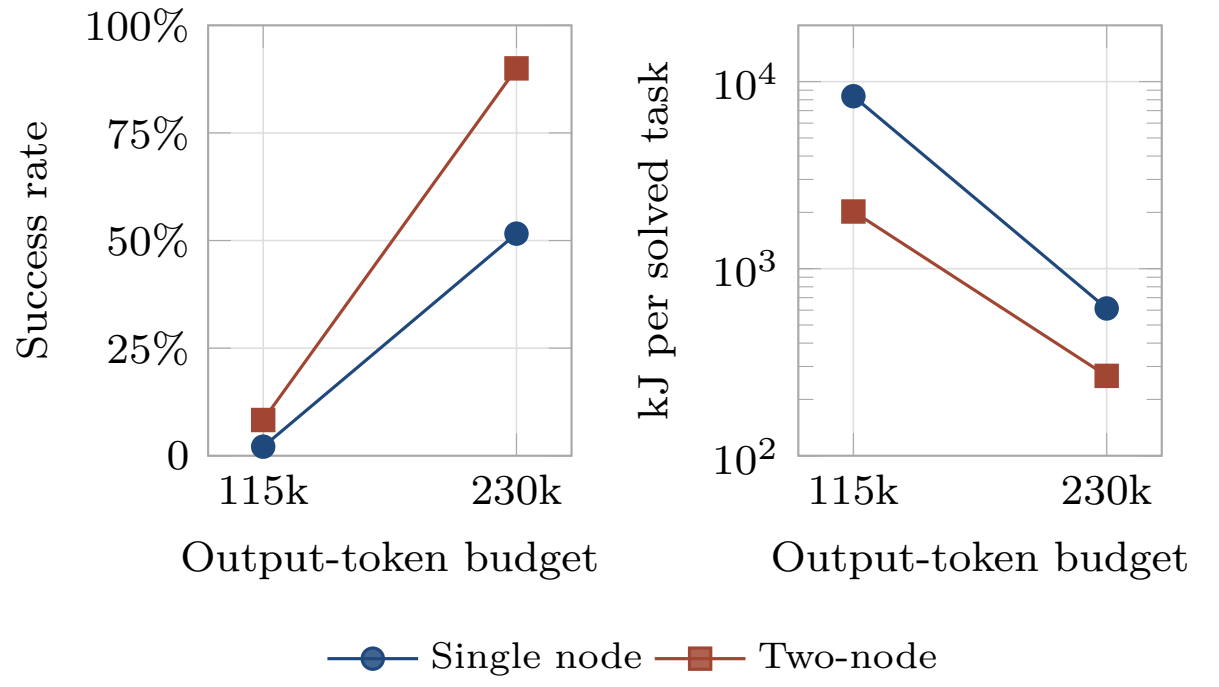}
\caption{Effect of doubling the output-token budget on construction tasks: success rate (left) and energy per solved task (right) for the single-node and two-node configurations. Single-node 230k uses the 62 runs with reasoning enabled; two-node 230k pools 10 runs including the control batch.}
\label{fig:budget}
\end{figure}

The token budget is the variable with the most direct effect on $p$. At the 115k output budget, Flash-Next passed 1 of 48 construction runs. Moreover, 26 of the 48 deliverables were byte-for-byte identical to the empty template model. In these runs, the agent had used up its budget before writing anything. The two-node model performed no better at this budget: 10 of its 11 failures ended at the cap. A stop at the cap does not by itself identify the cause, which may lie in the budget, in how the model works with the simulation tools, or in both. As shown in Fig.~\ref{fig:budget}, doubling the budget to 230k raised the single-node count to 32 of 62 ($P < 10^{-9}$) and the two-node count to 9 of 10 ($P < 0.001$), and cut the energy per solved task by a factor of eight to fourteen. A control batch with the task text still stating 115k but the enforced budget set at 230k passed 4 of 4, and on the two tasks concerned the count rose from 1 of 8 to 4 of 4 ($P = 0.013$). Therefore, the operative variable is the enforced budget, not the stated one. The budget notice tests Implication~6. With every tool result carrying the tokens used and remaining, the pass count on two construction tasks did not change significantly (12 of 16 against 14 of 24, $P = 0.33$). However, with the notice, runs stopped at the cap fell from 10 of 24 to 2 of 16, runs stopped with a passing model already on disk fell from 4 of 24 to none, and the energy per solved task fell by about 40\% (Section~\ref{sec:tradeoffs}). The notice supplied the one state variable that the loop had not fed back, and the agent used it to stop in time.

\subsection{Time}
\label{sec:timeres}

\begin{figure*}[!t]
\centering
\input{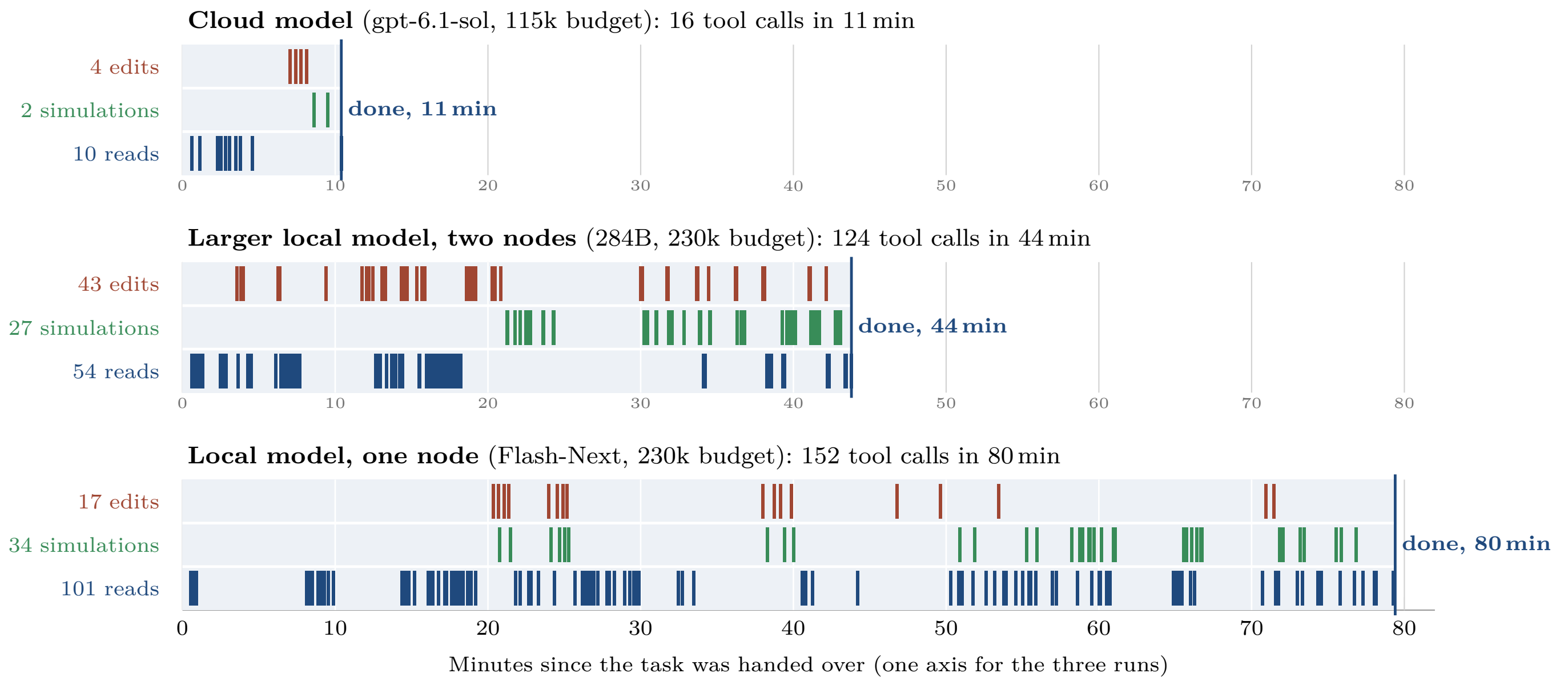}
\caption{One construction task, passed by three deployments, drawn on one time axis: the vertical grid and the tick values are the same for the three runs. Each tool call is a mark in the lane of its kind (edit: build or edit the model; simulate; read: read the model, results or files), and calls through generic code or shell tools are classified by the code they carry; the lane labels give the number of calls. The shaded band ends where the run was done, and the space between marks is model time.}
\label{fig:timeline}
\end{figure*}

Implication~7 splits the wall-clock time of a run into model time and tool time, and on the local device the model time dominates. In the held-out converter batches, the device was busy for 94\% of the run window. The model time is the output tokens over the decode rate. On the construction run of Table~\ref{tab:breakdown}, Flash-Next generated 22 output tokens per second of model time, compared with 55 for the cloud reference on the same task, and the two-node configuration decodes nearly twice as fast as one node. A passing construction run took 80 to 182\,min (median 121) on the local model against 10 to 49\,min (median 23) in the cloud. The local model produced more output tokens and produced them more slowly. Where the simulator is slow, tool time enters as well, in the form of the duty cycle shown in Section~\ref{sec:energy}. Fig.~\ref{fig:timeline} shows the tool calls of one construction task that all three deployments passed. The cloud model read the libraries it needed, built the model in four edits and verified it in two simulations within 11\,min. The two-node local model reached the same result in 44\,min by building in 43 small edits and simulating between them. The single-node model took 80\,min, most of it in model time between calls, with long stretches of reading files and models before its first edit.

\begin{table*}[!t]
\caption{Power System Set: Outcome (P pass, F fail), Wall-Clock Time (min), Simulations, Output Tokens (k) and Peak Context (k) per Attempt; Cloud Reference and Local Attempts 1 and 2}
\label{tab:pscadres}
\centering
\footnotesize
\setlength{\tabcolsep}{5pt}
\renewcommand{\arraystretch}{1.08}
\begin{tabular}{@{}l ccccc ccccc ccccc@{}}
\toprule
 & \multicolumn{5}{c}{Cloud reference} & \multicolumn{5}{c}{Local, attempt 1} & \multicolumn{5}{c}{Local, attempt 2} \\
\cmidrule(lr){2-6}\cmidrule(lr){7-11}\cmidrule(lr){12-16}
ID & & min & sims & tok & ctx & & min & sims & tok & ctx & & min & sims & tok & ctx \\
\midrule
S9 & P & 11 & 13 & 6 & 33 & P & 94 & 52 & 70 & 127 & P & 89 & 57 & 73 & 124 \\
S12 & P & 11 & 3 & 7 & 30 & P & 94 & 15 & 83 & 131 & P & 177 & 24 & 142 & 204 \\
S8 & P & 21 & 18 & 8 & 33 & P & 105 & 56 & 59 & 110 & P & 83 & 34 & 63 & 120 \\
S4 & P & 46 & 80 & 17 & 63 & P & 135 & 137 & 85 & 126 & P & 160 & 216 & 92 & 152 \\
S11 & P & 11 & 19 & 14 & 42 & P & 151 & 35 & 124 & 204 & P & 155 & 54 & 133 & 198 \\
S5 & P & 26 & 25 & 18 & 46 & P & 155 & 57 & 123 & 177 & P & 145 & 39 & 126 & 177 \\
S3 & P & 78 & 67 & 128 & 206 & F & 180 & 17 & 164 & 215 & F & 180 & 22 & 160 & 206 \\
S2 & P & 117 & 139 & 50 & 114 & F & 174 & 182 & 148 & 193 & P & 178 & 90 & 139 & 202 \\
S1 & P & 132 & 66 & 28 & 71 & F & 148 & 55 & 113 & 196 & P & 156 & 57 & 108 & 180 \\
S10 & P & 152 & 292 & 43 & 110 & P & 160 & 201 & 129 & 217 & P & 172 & 237 & 109 & 166 \\
S6 & P & 178 & 342 & 58 & 109 & P & 150 & 120 & 67 & 120 & P & 138 & 195 & 75 & 122 \\
X2 & P & 165 & 241 & 82 & 145 & P & 157 & 109 & 134 & 191 & P & 144 & 142 & 92 & 141 \\
S7 & P & 165 & 63 & 56 & 126 & F & 178 & 33 & 130 & 207 & P & 170 & 33 & 150 & 229 \\
\bottomrule
\end{tabular}

\end{table*}

As shown in Table~\ref{tab:pscadres}, the power system set shows the same pattern on a second simulator, with the cloud reference at its lowest tier. The first local attempts took a median of 151\,min and never less than 94, so they lie in a band between about 1.5\,h and the 3\,h limit. In contrast, the cloud reference took a median of 78\,min and as little as 11. Therefore, the ratio of the local time to the cloud time, from 13.7 on S11 to 0.84 on S6, follows the cloud time and does not mean that the local model is relatively faster on harder tasks. On the two tasks where the local attempt was shorter (S6, X2), the cloud model had used nearly the whole limit. The token counts explain that band. The local attempts produced 4.1 times the cloud's output tokens at the median task, with reasoning making up 73\% of the local output. In contrast, the number of simulations does not explain the difference. The first local attempts ran 1,069 in all against 1,368 for the cloud, fewer than the cloud on six tasks (S3, S1, S10, S6, X2, S7) and more on the other seven. The peak context of a single request exceeded 131k tokens on eight of thirteen local attempts. Therefore, with this model, tasks of this kind did not fit a 131k context window.

\subsection{Energy, Time and Token Costs per Solved Task}
\label{sec:four}

\begin{figure*}[!t]
\centering
\input{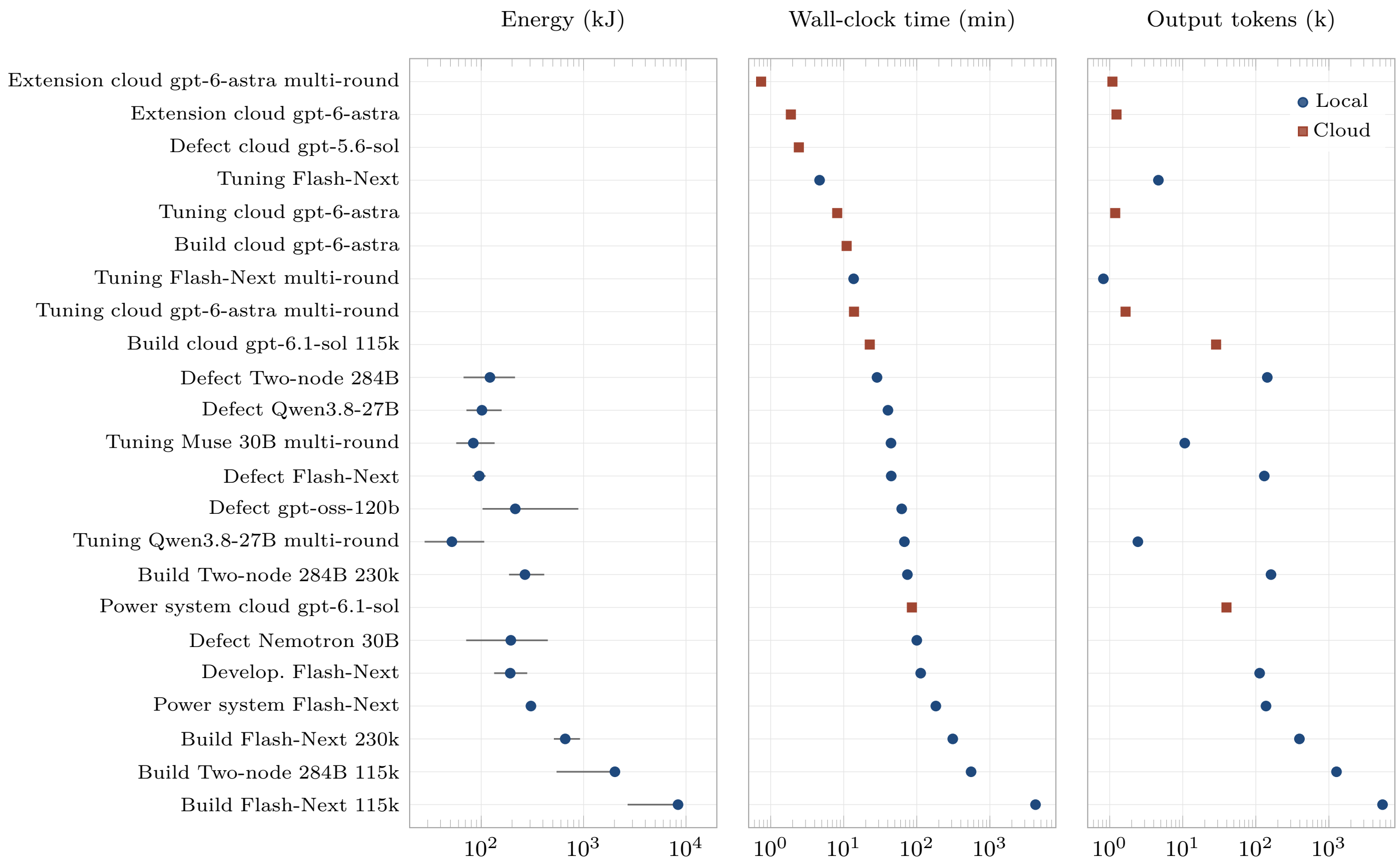}
\caption{The costs per solved task for each test condition (total over passes), for test conditions of at least ten runs, sorted by time: energy with bootstrap 95\% intervals conditional on at least one pass (local test conditions with a power record), wall-clock time and output tokens (local and cloud). Missing points are test conditions without that record. Passes over runs for each test condition are given in Fig.~\ref{fig:coverage}.}
\label{fig:eps}
\end{figure*}

\begin{figure}[!t]
\centering
\includegraphics{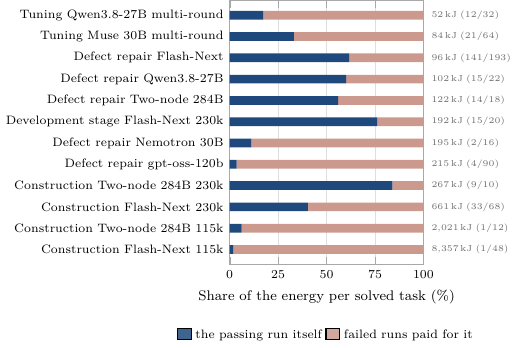}
\caption{Breakdown of the energy per solved task: for every test condition with at least ten runs and an energy record, the share of the energy per solved task that is the passing run itself (blue) and the share that is the failed runs incurred for each pass (red); the energy per solved task and the passes over runs are given at the bar end. Where the success rate is low, the failed runs dominate the cost.}
\label{fig:epsbreak}
\end{figure}

The cost per solved task is governed by the success rate, and the differences in the cost of a single run are secondary. Fig.~\ref{fig:eps} gives the energy, time and output tokens per solved task for every test condition of at least ten runs. Power is a rate, so it enters the cost of a solved task only through the energy. In Fig.~\ref{fig:epsbreak}, the energy is split into the passing run itself and the failed runs incurred for it, which is \eqref{eq:esuccess} evaluated for each test condition. The energy spans two orders of magnitude. It runs from about 50\,kJ for multi-round tuning to over 8,000\,kJ for construction at the 115k budget, and \eqref{eq:esuccess} accounts for the whole of it with the measured $p$ and $E_\mathrm{fail}$. For Flash-Next construction at 115k, $p = 1/48$ and a failing run is a full-budget run of about 167\,kJ, which gives $E_\mathrm{success} \approx 163 + 47 \times 167 \approx 8{,}000$\,kJ against 8,357\,kJ measured, about 51\,pu of $E_\mathrm{pass}$. Because the figure is the expression applied to the means of each test condition, this is a consistency check rather than a validation. The spread is explained by the success rate alone. Two test conditions test the corollary that decoding speed matters little when $p$ is small. gpt-oss-120b has the lowest-energy single run in the figure, 8.6\,kJ at the median. However, its energy per solved task is 215\,kJ, because it passed 8 of 111 defect runs and the energy of the other 103 is included. The two-node configuration decodes nearly twice as fast as the single node. However, on the defect tasks that both solved 10 of 10, it drew 1.15 times the energy for the same solved tasks, within the batch-to-batch drift of the plateau. Its lower energy per solved task in the construction test conditions comes with a different $p$, two devices against one, and a different model. A change of decode speed acts on $E_\mathrm{pass}$ and $E_\mathrm{fail}$ together and leaves the factor $(1-p)/p$ unchanged. Base-model scale affects both terms in the same direction. As shown in Table~\ref{tab:strata}, within one family and one model, a few hundred blocks against a three-thousand-block node gives 83\% against 57\% passes and a four-fold energy per solved task. Therefore, the pooled defect figure of 95.7\,kJ is a property of the mix of bases in the dataset.

Time and tokens per solved task follow the same expression and show the same spread. For Flash-Next construction at 115k, a solved task cost about 70\,h of wall-clock time and 5.4M output tokens, against 5.2\,h and 0.40M at 230k. The cloud reference needed 23\,min and 29k tokens per solved task on the same tasks. On the power system set, a solved task cost 182\,min, 138k output tokens and 306\,kJ locally, compared with 86\,min and 40k tokens in the cloud. The ranking of the test conditions is almost the same in all three units, because each is dominated by the same factor $(1-p)/p$. Between deployments, the time and token gaps are wider than the success-rate gap alone, because a local run on the single device used here is also slower and longer.


\subsection{Comparison with a Cloud Model}
\label{sec:cloud}

On the three construction tasks above, the cloud model (gpt-6.1-sol at its highest reasoning tier) passed 12 of 12 at the 115k budget while using only 12\% to 49\% of it. The budget that stopped the local model before it wrote anything was more than twice what the cloud model needed. A control batch at the model's lowest reasoning tier, run after the dataset freeze and otherwise identical, passed the same tasks 12 of 12 in 3--9\,min per run. Therefore, the reasoning tier does not account for the gap, and against the local 1 of 48 the contrast is $P < 10^{-8}$ at either tier. The cloud model is also affected by the reasoning effort. On a separate set of six data-centre converter tasks, a low and a high reasoning tier passed the same number of runs, 6 of 12 each, and the high tier produced 2.2 times the output tokens. No efficiency comparison is drawn, because the cloud has no energy record.\footnote{The comparison is between deployments and not between equal models: the open-weight models that can be served locally are generally placed some months behind the closed frontier models \cite{R34_epoch_eci,R35_epoch_lag}, so part of the gap reflects the model rather than its location. Serving the same open-weight model on rented data-centre accelerators would separate the two effects. This was not measured here, and a higher decode rate at unchanged model output is to be expected.} The comparison shows that doubling the local budget provides the additional time that the cloud model did not require, and that reasoning effort has a cost in both deployments.

\subsection{Effect of Interventions}
\label{sec:tradeoffs}

\begin{figure}[!t]
\centering
\includegraphics{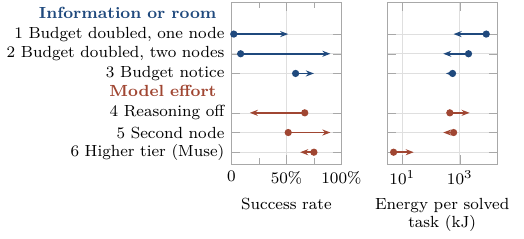}
\caption{Interventions, one per row: success rate (left) and energy per solved task (right, logarithmic) before (dot) and after (arrowhead). Interventions that give the agent information or room (blue) raise the success rate and lower the energy per solved task; interventions on model effort (red) do not in these test conditions. Numbering as in the text: 1, 2 budget 115k to 230k; 3 budget notice; 4 reasoning enabled to disabled (230k); 5 one to two nodes (230k); 6 reasoning tier low to high (Muse, tuning). Measured feedback (3/160 to 15/16) is not drawn, because its single-submission test condition has no energy record.}
\label{fig:levers}
\end{figure}

The choices listed in Section~\ref{sec:intro} can be read through \eqref{eq:esuccess}, i.e., an intervention lowers the cost per solved task only if it raises $p$ faster than it raises $E_\mathrm{fail}$. Interventions that change what the loop gives the model (a measured result, a budget sufficient to finish, an indication of the remaining budget) act directly on $p$. In contrast, interventions that increase the computation on each turn (reasoning, a higher reasoning tier, a larger model on more hardware, more documentation in the context) act mainly on $E_\mathrm{fail}$, and on $p$ only if the additional computation yields what the task requires. Fig.~\ref{fig:levers} plots the interventions that have both a success rate and an energy record before and after, while the others are given in the text. The three interventions that gave the agent more information or a larger budget lowered the cost. In contrast, the interventions on model effort did not act in one direction. A higher reasoning tier, a second node and added documentation did not lower the cost in the test conditions measured, whereas removing reasoning altogether raised it. No effort contrast reaches $P < 0.05$. A third class, device settings, lowered energy without acting on the task at all.

\begin{table}[!t]
\caption{Device Settings That Lower Energy Without Acting on the Task. Calibrated Separately on Fixed Workloads on the GB10 Device; No Pass Rate Was Measured}
\label{tab:settings}
\centering\footnotesize
\renewcommand{\arraystretch}{1.1}
\begin{tabular}{@{}>{\raggedright\arraybackslash}p{2.5cm}>{\raggedright\arraybackslash}p{5.5cm}@{}}
\toprule
Setting & Effect on the fixed workload \\
\midrule
GPU clock locked at half its default & Task energy $-66$\%; wall-clock time essentially unchanged \\
Speculative decoding & Energy per output token $-32$\% \\
Four concurrent agents & Energy per output token $-60$\%: power $+1.6$\,W, throughput $\times 2.6$ \\
Model kept resident & Break-even against a cold start at about 18\,min of idle \\
\bottomrule
\end{tabular}
\end{table}

Measured feedback is the clearest case of giving the agent more information. A single submission from Flash-Next on a tuning task passed 3 of 160 runs. Once the agent saw the simulated result after each proposal and changed one quantity per round, the count rose to 15 of 16, and on a second task from 1 of 32 to 31 of 32. However, the effect is confined to tasks that are feasible as posed. On infeasible ones, the agent still proposed settings in most runs, and only a final confirmation round recovered two of sixteen.

The effect of the effort interventions can be explained by the loop, and it is not always in the same direction. Without reasoning, the agent acted and observed turn after turn, making five times as many tool calls per run. As a result, the context grew faster, four of six runs reached the cap, and the energy per solved task rose five-fold (Fig.~\ref{fig:levers}, row 4). However, on a tuning task with reasoning enabled, eight runs spent 4.4\,h of reasoning for three passes, against about 2\,min per success with reasoning disabled. The right amount of reasoning depends on the task. Raising Muse's reasoning tier produced more tokens per turn without new information in the context, and the pass count fell while the energy per solved task rose five-fold (row 6). A second node decodes the same tokens faster at about twice the power, so the energy per solved task cannot change unless $p$ does, and on the saturated defect tasks it did not. The pooled construction advantage of the two-node configuration (row 5) cannot be separated from its different model. Injecting the tool manual into the context produced no passes in 16 runs and reduced gpt-oss from 1 of 18 to 0 of 18 at unchanged power, because the manual is re-sent every turn and displaces working context. The device settings of Table~\ref{tab:settings} act on the plateau and on the time per token rather than on the task. A locked clock has saved up to 32\% of decode energy on data-centre GPUs as well, where a sweep of the clock places the minimum of energy per token below the default \cite{R42_ma_capping,R53_kim_b200,R72_powertrace}. These settings scale $E_\mathrm{pass}$ and $E_\mathrm{fail}$ together and have no effect on a task the agent cannot solve.

\section{Discussion}
\label{sec:discussion}

This section discusses what the measurements mean for an engineering group that considers a local agent: how the costs should be read, which tasks suit a local deployment, where the cost lies, how a deployment is sized, what the evaluation procedure requires, and how this work can be extended.

\subsection{How to Read the Costs}
\label{sec:factors}

A success rate or a cost reported in this paper is a property of a configuration, not of a model. It should be read like a datasheet value, which is meaningful only together with its test conditions. The configuration comprises the choices of Fig.~\ref{fig:coupling} together with the task, the verification and the hardware, and several of these factors changed the outcome more than the choice of model did. For example, the output-token budget alone changed the construction result from 1/48 to 32/62. Enabling reasoning changed the success rate of one model on the same family from 17\% to 67\%. Within one family and one model, the scale of the base model alone changed the success rate from 83\% to 57\% and the energy per solved task four-fold. On the hardware side, clock locking changed the energy by 66\% at unchanged wall-clock time, and the decode plateau drifted by about 25\% across batches. Therefore, a comparison between two models or two deployments is meaningful only when these factors are held equal, which is the purpose of the procedure of Section~\ref{sec:framework}.

\subsection{Implications for Local Deployment}
\label{sec:decision}

Local deployment is chosen for reasons such as data sovereignty and on-premises simulator licensing. The question that remains is which tasks are suited to it, and \eqref{eq:esuccess} gives the criterion. A task is viable locally when $p$ is high enough that the energy, the time and the tokens of a solved task, each scaled by $(1-p)/p$, are acceptable together. Neither a faster device nor a cheaper model per run changes that factor, as the two-node and gpt-oss test conditions showed (Section~\ref{sec:four}).

In the configuration measured here, the criterion separated the tasks by where they start. The tasks that start from an existing model met it at the costs of Section~\ref{sec:four}: 1) diagnosing and repairing a model; 2) carrying out a step of an existing design; 3) tuning a controller against the simulated result of each proposal; and 4) a power system study that returns a setting or a number from a given model. In contrast, building a converter model from a specification did not meet it at the smaller token budget, and at the doubled budget it cost hours per solved task against minutes in the cloud (Sections~\ref{sec:four} and~\ref{sec:cloud}). Where this line falls is a property of the model, the budget and the device, and it will move as they change. However, the criterion, and the need to measure $p$ on the tasks concerned, will not.

The token budget has a threshold. Below it, only the runs that finish early can pass, so $p$ is low and every cost per solved task is high. A group can locate this threshold for its own task mix by measuring $p$ on a small number of runs at two budgets. On the power system set, both deployments reached the same results wherever both passed. On this set, the further requirements of local deployment were a context window above 131k tokens for the model used and the effort of maintaining the safeguards, and its benefit is that the model, the study files and the transcripts remain on site.

\subsection{Where the Cost Lies and What Lowers It}
\label{sec:wherecost}

At the scale of this study, a local agent is a small electrical load. The device consumes tens of watts, and a solved defect task costs a few tens of kilojoules. Therefore, electricity is the smallest item in a cost model for engineering use, and wall-clock time is the next (Section~\ref{sec:timeres}). Both were measured on a single small inference device, chosen as the minimum on which the study could be made, not as a deployment. A deployment that has to be faster can run several nodes in parallel or a device with more compute and memory bandwidth (Section~\ref{sec:deploy}). In that case, its time per solved task falls, its electricity becomes a significant item rather than the smallest one, and the depreciation of the hardware is likely to be the more pressing cost.

The largest item is labour, which no meter records. An independent verification script and fixed criteria have to be written and calibrated for each new task type, the runs have to be isolated so that one run cannot read another's answers, and the records have to be complete enough for a result to be audited. All of this had to be completed before the first comparable result was obtained.

The metered costs are reduced first by improving what the agent observes, and only then by the size or the reasoning effort of the model. Two measures were the most effective. The first is to return the simulated result of each proposal to the agent. Its detail is set by the tool designer rather than by the model, and it is traded against prefill, since every byte returned is re-sent on every later turn. The second is the budget notice, which removed the failures that an unseen budget produces.

\subsection{Sizing a Deployment from These Measurements}
\label{sec:sizing}

With the constants that the measurements supply, the relations of Section~\ref{sec:deploy} give four sizing rules. Appendix~\ref{app:sizing} gives the calculation for the main model of this study in detail.

First, the model fixes the memory class of the device before any task is run. In NVFP4, the weights of the main model occupy about 126\,GB, so by \eqref{eq:capacity} the device class is 128\,GB. The full 262k context window is served on that device only because the embedding table of the model is mapped from solid-state storage and not held in memory.

Second, the memory bandwidth bounds the decode rate, and the attained rate lies well below the bound. On the measured device, the main model attained a fraction of about 0.2 to 0.3 of the bound of \eqref{eq:rdec}, and a dense model about 0.6. Therefore, a rate estimated for another device from its bandwidth assumes the same engine and the same attained fraction, and it is not a measurement. The rate also sets the waiting time. The construction run of Table~\ref{tab:breakdown} would need about 30 tokens/s to return within an hour, and about 90 tokens/s, close to four times the device used here, to return within 20 minutes, the order of the cloud median. By \eqref{eq:esuccess}, the rate that matters is that of a solved task, so the requirement scales with $1/p$ as well.

Third, prefix caching in the serving engine is a requirement rather than an option. Agent runs carried contexts above $10^5$ tokens, and at the cache hit rate of Section~\ref{sec:tokens}, i.e., 94.9\% of the input, an engine without prefix caching would do about twenty times the prefill work.

Fourth, the idle power acts as a no-load loss. The energy per output token during decode is about 1.4\,J for the main model, of which about 1.0\,J is the increment above the idle power (Section~\ref{sec:energy}). The remainder is the idle power, about 0.37\,pu, which is also consumed in every hour that the device stays loaded without decoding (Section~\ref{sec:power}).

The serving engine and concurrency have effects comparable to that of the device class. However, each of them gives a new configuration whose $p$ has to be measured again. For example, a published engine reaches about 1.7 times the rate measured here for the same model on the same device \cite{R59_strata}, but on short answers, at shorter contexts and with speculative decoding, which was not used in the evaluated runs. Concurrency raises the throughput of a set of short tasks and does not shorten the completion time of a long one. On agent tasks whose context had grown past $10^5$ tokens, five concurrent streams together decoded at about the rate of a single stream.

\subsection{The Evaluation Procedure}
\label{sec:durability}

A cost per solved task is only as reliable as the verification on which the count of solved tasks rests. Three lessons from applying the procedure are summarized here.

First, the evidence has to be independent of the agent. Three independent records are kept for every run, i.e., the executor's log of what the model proposed, the tool's record of what was executed, and the independent recomputation of the result. None substitutes for another. Rules that matter are enforced in the tool rather than requested in the prompt, because prompt instructions alone did not prevent an agent from reading stale answers. In addition, summary metrics are recomputed from full recordings. For example, a distortion computed on integer harmonics alone gave a few percent on a model whose full spectrum showed about 41\% distortion.

Second, a cost has to be reported together with its conditions. A cost per solved task should be reported with $p$, the failure cost, the token budget, the base-model scale and the invalid-run policy. A plateau should be distinguished from a window mean, and a token total should state what share was served from cache.

Third, the procedure keeps the costs meaningful as models improve. At its lowest reasoning tier, the cloud reference already passed every converter construction task and every power system task, so neither set measures that model's capability any longer. However, both still measure what a solved task costs, and that cost does not vanish with capability. The same model's token count doubled at a higher tier for the same pass count. When a set saturates, its difficulty can be raised, and the result never relies on the agent's self-report. The procedure also audits the task author as well as the agent. Defects on the author's side of the power system set were corrected because an agent's result disagreed with them.

\subsection{Further Considerations}
\label{sec:limitations}

The framework is intended to be reused and extended, and several directions follow directly from this study. Appendix~\ref{app:scope} states the scope of the present evidence in detail.

The first direction is breadth. In this study, one main model and one device type carried most of the local runs. Because the procedure and the relations do not depend on them, the same tests can be repeated on other models, devices and executors, and each repetition adds a configuration to the comparison. With more runs in each test condition, the intervals narrow and smaller effects, e.g., those of the reasoning effort, can be resolved. The executor is one such factor. The one used here is the cloud vendor's own, and running the local models under other executors would separate the fit of the executor from the capability of the model \cite{R56_openjarvis}.

The second direction is the measurement boundary. Energy was measured here as GPU-domain power integrated over the run window. Extending the boundary to the AC input of the device and to the agent host would give the energy of the complete installation. In addition, an energy record of a cloud deployment would allow the two deployments to be compared in energy as well as in tokens and time. On the hardware side, the same measurement on discrete-GPU hardware would show how far the plateau carries over to devices with different idle and prefill behaviour, and running concurrent agents on one inference server would show how they share it.

The third direction is the workflow. The power system set was run with one cloud attempt and two local attempts per task, and repeating the attempts would turn its counts into success rates. Moreover, all runs measured here were unattended. A supervised workflow, in which the engineer takes the last step that the local model missed, is the natural next case, and its cost in engineer time can then be set against the unattended runs measured here.

Beyond these directions, several topics appear promising for future research: 1) extending the agent loop from simulation to hardware-in-the-loop and laboratory test benches; 2) compact open-weight models adapted to power electronics tasks, which fit smaller inference devices; 3) simulation tools designed for agents, which return concise and verifiable results; 4) load models and scheduling of groups of agents as electrical loads; and 5) shared task sets and reporting conventions for the cost per solved task. As models and devices continue to improve, it is hoped that such efforts can make local agents a practical and measurable part of the engineering workflow in power electronics.

\section{Conclusion}
\label{sec:conclusion}

In this paper, a systematic evaluation framework is presented for agentic AI that is deployed locally and drives power electronics simulation. Its purpose is to prepare the ground for practical application, i.e., to show what such a deployment involves, how it can be measured objectively and fairly, and what it costs. The work is summarized in three aspects.

\begin{enumerate}
\item \textit{Deployment:} The constraints of a local deployment are analysed from the structure of the agent loop. Expressions are derived for the power, energy, tokens and time on the inference device, and for the expected cost of a task that is run until it succeeds. These relations show how the choices of a deployment are coupled, and they give a basis for sizing the inference device.
\item \textit{Measurement:} A unified procedure is established, in which various agents are tested on the same tasks under identical conditions, every result is verified independently of the agent's own report, and the four quantities are recorded for every run. The procedure does not depend on a particular simulator, inference device or model, and it is applied here to tasks from the converter level to the system level.
\item \textit{Cost:} In power, energy, tokens and time alike, the cost per solved task is determined mainly by the success rate on that task rather than by the cost of a single run. Therefore, whether a task is suited to a local agent has to be decided by measuring the success rate on the tasks concerned. Moreover, two measures reduce this cost most effectively: feeding the simulated result of each proposal back to the agent, and giving it a sufficient output-token budget together with a notice of how much of the budget remains.
\end{enumerate}

The capability of agentic AI is changing quickly. The boundaries reported in this paper belong to the models, the budgets and the device measured, and they will move as these change. However, the framework that locates them will not. It is hoped that this paper can serve as a preparatory study for the practical use of agentic AI in power electronics, and as a template that other groups can repeat on their own tasks, devices and models.

\appendices
\section{Models}
\label{app:figs}

This appendix shows the models the agents worked on and two of the models they submitted.

Fig.~\ref{fig:app_simulink} shows two buck converters built from an empty template model in the construction family of Section~\ref{sec:cost}, one by the cloud model and one by the local single-node model. Both passed verification against the specification. The cloud model organized its circuit into subsystems, while the local model placed all blocks on one level, with a synchronous switch pair, a soft-start reference and an inner current loop. Note that layout is not scored. The agents never see the power system models themselves. Instead, each is accessed through a simulation service that runs the frozen model on a single licensed Windows session and returns the requested waveforms as a file.

\begin{figure}[!t]
\centering
\input{figures/fig_app_simulink}
\caption{Two buck converters built from an empty template model in the construction family, both verified as passing: (a) the power-stage subsystem of the cloud model's build; (b) the local single-node model's build of the same task, with a synchronous power stage and cascaded voltage and current loops drawn on one level.}
\label{fig:app_simulink}
\end{figure}

\begin{figure*}[!t]
\centering
\input{figures/fig_app_bases}
\caption{Base models of the converter set, as frozen for the batches. The agent reaches them only through the tool interface and never sees these diagrams.}
\label{fig:app_bases}
\end{figure*}

\begin{figure*}[!t]
\centering
\input{figures/fig_app_pscad}
\caption{PSCAD/EMTDC main canvases of seven of the thirteen system-level test beds and one module, as frozen for the batches. The agents never see these canvases; each test bed is reached through the simulation service.}
\label{fig:app_pscad}
\end{figure*}

Fig.~\ref{fig:app_bases} shows six of the base models of the converter set, from a buck converter of a few hundred blocks to the grid-tied microgrid node of about three thousand, and Fig.~\ref{fig:app_pscad} shows the main canvases of the system-level test beds.

\section{Power System Set: Attempts in Detail}
\label{app:pscad}

This appendix gives the account of the power system attempts that Section~\ref{sec:capability} summarizes, and the thirteen tasks are listed in Table~\ref{tab:pscad} with the identifiers used throughout the paper.

\begin{table*}[!t]
\caption{Power System Set: Reference Model and Deliverable of the Thirteen Tasks}
\label{tab:pscad}
\centering
\footnotesize
\renewcommand{\arraystretch}{1.05}\setlength{\tabcolsep}{6pt}
\begin{tabular}{@{}llll@{}}
\toprule
ID & Layer & Reference model & Deliverable \\
\midrule
S1 & System & IEEE 39-bus system & Critical clearing times at several buses \\
S2 & System & Single machine, infinite bus, with PSS & PSS parameters for damping and gain margin over several operating points \\
S3 & System & 132/33\,kV OLTC with a feeder & Tap-changer control settings for voltage regulation \\
S4 & EMT & 400\,kV line, statistical switching & Pre-insertion resistor from a catalogue \\
S5 & EMT & 230\,kV capacitor banks & Inrush reactor from a catalogue \\
S6 & EMT & 230/13.2\,kV transformer on an open line & Fault-record diagnosis and a ferroresonance damping resistor \\
X2 & EMT & 11/0.6\,kV transformer with remanent flux & Controlled-energization closing rule \\
S7 & IBR & 125\,MVA PV plant, 230\,kV & Type-test result table and ride-through settings \\
S8 & IBR & Grid-forming PV inverter, 345\,kV & Minimum short-circuit ratio for energization and operation \\
S9 & IBR & 100\,MVA grid-following converter, 220\,kV & Minimum short-circuit ratio and the oscillation frequency beyond it \\
S10 & HVDC & CIGRE HVDC benchmark & Commutation-failure boundary in fault resistance \\
S11 & HVDC & Hybrid HVDC breaker test bench & Current-limiting reactor and arrester rating from catalogues \\
S12 & HVDC & MMC link with a long DC line & Fault location and instant from terminal recordings \\
\bottomrule
\end{tabular}
\end{table*}

\begin{figure*}[!t]
\centering
\input{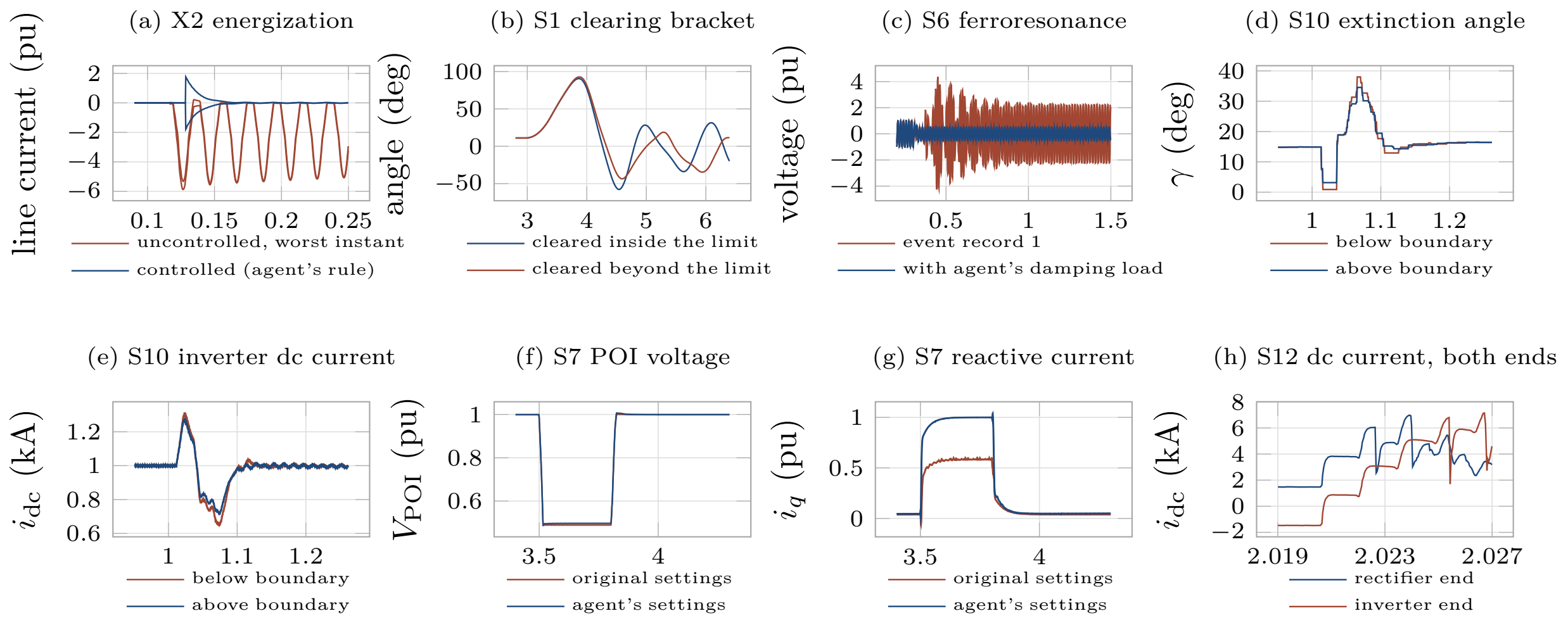}
\caption{Waveforms from runs the local agent made during its second attempts, with the task's event record in (c). Time axes in seconds. Each trace is drawn as the per-bin minimum and maximum of the returned samples, so switching transients survive the decimation. (a) X2, line current at transformer energization, uncontrolled against the agent's controlled close. (b) S1, rotor angle difference for a fault cleared at two durations either side of the stability limit. (c) S6, bus voltage in the recorded ferroresonance event and in the agent's verification run with its damping load. (d), (e) S10, extinction angle and dc current of the LCC-HVDC inverter for a fault either side of the commutation-failure boundary. (f), (g) S7, POI voltage and reactive current of the solar plant in a fault-ride-through test, with the original settings and with the agent's settings. (h) S12, dc line current at both terminals in the agent's calibration run.}
\label{fig:app_waves}
\end{figure*}

Three cloud tasks were re-run on revised task packages after the first pass exposed defects in the tasks, and the local attempts were run in order of expected duration. The first four local attempts ran with a 131,072-token server context, which stopped one attempt (S11) without a deliverable. That attempt was recorded as invalid, and the context was raised to 262,144 for all later attempts. The rules for reading each result were written before the batch, as in Section~\ref{sec:fairness}. Five submissions, three local and two cloud, were made after the agent itself had measured a failing case, and on three of them the undisclosed cases did not contain the failing condition the agent had found. The outcome, time, simulations, tokens and context of every attempt are given in Table~\ref{tab:pscadres} in Section~\ref{sec:timeres}.

On two of the first local attempts, the local result was the closer one (S10, boundaries within 2\% of the truth against 2--4\% for the cloud; S12, zero location error against up to 2.1\,km).

The second local attempts, run in the same order under the same conditions, passed twelve of thirteen. Ten outcomes repeated the first attempt, with the same result wherever both passed. S2, S1 and S7 passed on the second attempt after failing on the first. S3 reached the limit twice, the second time with a setting that the undisclosed profiles of the verification script accepted but without the report the task requires. Time changed by less than 10\% on ten tasks. The one large change is S12: the second attempt reached the same zero-error result but took 177\,min against 94, spending the difference on a longer calibration and on bit-exact verification runs before submitting with 2.5\,min remaining. The three reversals were not repeats with better luck. On S2, the second attempt treated the operating point that its first submission had not met as a hard constraint and added a second compensation stage. On S1, the second attempt recognized a non-monotone stability boundary and bisected on the outcome instead of stopping at the first loss of synchronism. On S7, the second attempt isolated the plant-controller windup with eight interim runs before modifying that loop, and the undisclosed case that had failed passed. Thus, the same model on the same task can differ by a factor of two in time, and in outcome, while the band of 1.5 to 3\,h does not change.

Fig.~\ref{fig:app_waves} shows waveforms from the second local attempts. They are the kind of result the agent reads to choose its next action, and the kind the independent verification script re-measures.

The four failures of the first local attempts are of three kinds. One is time. On S3, the local model identified the substation from 35 open-loop runs, built its own tap-changer simulator and searched for settings inside it. When that simulator reported every candidate as infeasible, the model did not go back to PSCAD, and it reached the limit with no deliverable. Two are judgement at submission. On S2, the model measured a damping ratio just below the required value at one operating point, recorded it in its report, and submitted the setting as a balance between conflicting networks. On S7, its own validation of 22 cases showed one failing by 0.017\,pu, and it submitted the settings anyway. The settings then failed one of the undisclosed cases. The fourth is one step short. On S1, the model found, at three buses, the shortest fault duration at which the system lost synchronism several swings after clearing, reported it as the critical clearing time, and did not test longer durations. The cloud model met the same conflicts on S2 and S7 and chose a second compensation stage and a looser threshold, respectively.

Four defects in the tasks and two disturbances were found during the batches and handled under the rules fixed beforehand, without re-judging any attempt. The cloud screening exposed two of the defects (S5, S3), and both packs were revised and re-run. The local attempt on S1 exposed a definition that did not anticipate the system's behaviour. The definition was made precise, the original reference value was retained, and the results of both deployments were unchanged. On S7, the cloud model pointed out an inconsistency in the task statement. A fifth defect was found by an agent rather than by the procedure. Specifically, the cloud report on the tap-changer task constructed a case inside the stated conditions on which its own submitted setting violated a requirement, and said so. After the batch, the local second-attempt setting was found to fail the same case, although both settings had passed all undisclosed cases. Both results stand under the rules fixed beforehand, and the task is revised for later use. The two disturbances were a 12\,s stop of the simulation service, and an author-side computation that entered the shared queue during the local batch and delayed two attempts by about 50 and 80\,min. Both delayed attempts passed and are recorded with an annotation. The procedure checked the task author as well as the agents, a property Section~\ref{sec:durability} returns to.

\section{Details for the Discussion}
\label{app:discussion}

This appendix gives the detailed calculation behind Section~\ref{sec:sizing} and the scope of the evidence referred to in Section~\ref{sec:limitations}.

\subsection{Sizing Calculation for the Main Model}
\label{app:sizing}

The relations of Section~\ref{sec:deploy} are applied here to the main model of this study, with the constants the measurements supply. Flash-Next has 176B parameters, i.e., a 125B main model, of which 6B are active per token, and a 51B n-gram embedding table that is only looked up. In NVFP4, its weights occupy about 126\,GB, and each output token reads about 3.4\,GB. Therefore, by \eqref{eq:capacity}, the model fixes the device class at 128\,GB before any task is run. It serves its full 262k context window on that device because the embedding table, which is only looked up, is mapped from solid-state storage and not held in memory. Thus, about 76\,GiB of main model stays resident, which leaves room for the key--value cache of a full window. For the devices of Table~\ref{tab:devices} that hold the model in their own memory, the bound of \eqref{eq:rdec} with $\eta = 1$ is about 26 tokens/s for a desktop processor, 75 for the Ryzen AI Max+ 395, 80 for the GB10 device, 181 for an EPYC 9005 and 353 for the M5 Ultra. On the GB10 device, the dense 27B model decoded at 10.3 tokens/s against a bound of about 17, and Flash-Next at about 24 tokens/s, the value fitted on the held-out batches, against a bound of about 80. These are attained fractions $\eta$ of about 0.6 and 0.3, respectively. Single runs decoded between about 16 and 24 tokens/s, depending on the server build and on the share of reasoning tokens. Therefore, the Flash-Next fraction lies between about 0.2 and 0.3. At a fraction of 0.3, these bounds become estimated rates of about 8, 22, 54 and 105 tokens/s for the devices other than the one measured. It should be noted that they assume the same engine and the same attained fraction, which can differ on another device and on a processor in particular, and are not measurements. The device was busy for 0.94 of the window in the held-out batches. Therefore, the construction run of Table~\ref{tab:breakdown}, 102k output tokens in 80\,min, would need about 30 tokens/s to return within an hour. To return within 20 minutes, the order of the cloud median, it would need about 90 tokens/s, close to four times the device used here. By \eqref{eq:esuccess}, the rate that matters is that of a solved task, so it scales with $1/p$ as well. Agent runs carried contexts above $10^5$ tokens, and at the cache hit rate of Section~\ref{sec:tokens}, i.e., 94.9\% of the input, an engine without prefix caching would do about twenty times the prefill work. The energy per output token during decode is about 1.4\,J for Flash-Next (a window-mean power of 33.7\,W at about 24 tokens/s). About 1.0\,J of it is the increment above the idle power (Section~\ref{sec:energy}). The remainder is the idle power, about 0.37\,pu, which is also added to every hour that the device stays loaded without decoding (Section~\ref{sec:power}).

The serving engine has an effect comparable to that of the device class. Strata reports 40 to 43 tokens/s for Flash-Next in NVFP4 on the same device \cite{R59_strata}, about 1.7 times the rate measured here. However, the reported rate is on short answers, at contexts up to 32k and with speculative decoding, which on this device lowered the energy per token by 32\% and was not used in the evaluated runs. A deployment on such an engine is a new configuration whose $p$ has to be measured again. Concurrency behaves the same way. Four concurrent streams lowered the energy per output token by 60\% on a fixed workload, and at a 19k-token context five streams together decoded 68 tokens/s, 2.9 times one. In contrast, on agent tasks whose context had grown past $10^5$ tokens, five concurrent streams decoded about 4.4 tokens/s each, together about the 21 tokens/s of a single stream. Concurrency raises the throughput of a set of short tasks and does not shorten the completion time of a long one.

\subsection{Scope of the Evidence}
\label{app:scope}

The scope of the evidence is as follows. One main model and one device type carry most of the local runs, while the other models appear in one to six test conditions each. Energy is GPU-domain power integrated over the run window and excludes the MATLAB host and the chassis. Test conditions hold between 2 and 581 valid runs, and the intervals in Fig.~\ref{fig:eps} are wide where test conditions contain few runs. The 789 Flash-Next tuning runs have no energy record. Moreover, single- and two-node energies are not a like-for-like efficiency comparison, and no cloud energy was measured. Success rates are counts under the stated conditions and not capability values. Two pools of the same test condition appear, i.e., construction at 230k on the single node as 33/68 in Fig.~\ref{fig:eps} (all runs) and 32/62 in Section~\ref{sec:tokens} (reasoning enabled), and the text states which. The exact tests are stratified by task but still treat repeated runs of one task as independent, so the $P$ values are indicative. No contrast between effort settings in Section~\ref{sec:tradeoffs} reaches $P < 0.05$. Moreover, the before and after test conditions of the budget notice and of the Muse reasoning tier were run in different batches, between which the decode plateau drifts by about 25\%. The executor is the cloud vendor's own and may suit the cloud reference better than it suits the local models, so its fit is confounded with capability \cite{R56_openjarvis}. The power system set has fewer runs still: one cloud attempt per task at the lowest reasoning tier, two local attempts, no power record for the first local pass, and tasks authored by the authors and corrected during the batches as described. Its counts show what one attempt did, not a rate. Finally, \eqref{eq:esuccess} is applied with the means of each test condition. It is an accounting identity, not a fitted model. Five questions remain open: 1) whether the plateau holds on discrete-GPU hardware with different idle and prefill behaviour; 2) whether the ratio of failure to success cost holds for a model generation that fails faster; 3) how concurrent agents sharing one inference server share the plateau; 4) what a cloud energy record would show if one existed; and 5) what a supervised workflow, in which the engineer takes the last step the local model missed, costs in engineer time against the unattended runs measured here.


\end{document}